%% file: paper.tex
\documentclass{article}
\PassOptionsToPackage{table}{xcolor}
\usepackage{iclr2027_conference,times}
\iclrfinalcopy
\usepackage[utf8]{inputenc}
\usepackage[T1]{fontenc}
\usepackage{amsmath,amssymb,amsthm}
\usepackage[table]{xcolor}
\usepackage{booktabs,graphicx,microtype,multirow}
\usepackage{subcaption}
\usepackage{listings}
\definecolor{codekeyword}{HTML}{345C9C}
\definecolor{codetype}{HTML}{765A91}
\definecolor{codestring}{HTML}{39764B}
\lstdefinelanguage{Lean}{
  morekeywords={def,theorem,by,unfold,omega,structure,where,let,fun,match,with,if,then,else},
  morekeywords=[2]{Data,State,Nat,Prop,Bool,Task,Heuristic},
  morestring=[b]",morecomment=[l]{--}}
\lstdefinelanguage{PDDL}{
  sensitive=false,alsoletter={:,-},
  morekeywords={define,domain,problem,and,not,:requirements,:strips,:typing,
    :types,:predicates,:action,:parameters,:precondition,:effect,:domain,
    :objects,:init,:goal},
  morecomment=[l]{;}}
\usepackage{hyperref}
\usepackage{url}
\hypersetup{hidelinks,pdftitle={LeanPlan: Optimal Planning with LLM-Generated Heuristics and Admissibility Proofs},
  pdfauthor={}}
\input{commands}

\usepackage{todonotes}
\usepackage[disable]{comment-macros} %
\usepackage{tikz}
\usetikzlibrary{arrows.meta}
\definecolor{cText}{HTML}{1D1A14}
\definecolor{cAccent}{HTML}{B4471F}
\definecolor{cGreen}{HTML}{1F4A3A}
\definecolor{cRed}{HTML}{8F2118}
\tikzset{
  stage/.style={draw=cText, line width=0.9pt, rounded corners=3pt, fill=white,
                align=center, inner sep=5pt, font=\footnotesize,
                minimum height=1.05cm},
  stageA/.style={stage, draw=cAccent, fill=cAccent!8},
  stageG/.style={stage, draw=cGreen, fill=cGreen!8},
  ar/.style={-{Latex[length=2.2mm]}, line width=0.9pt, draw=cText},
  arG/.style={ar, draw=cGreen, line width=1.1pt},
  arR/.style={ar, draw=cRed, line width=1.1pt}}

\title{LeanPlan: Optimal Planning with LLM-Generated Heuristics and Admissibility Proofs}

\author{Andr\'{e} G.\ Pereira\\
Federal University of\\
Rio Grande do Sul\\
Brazil
\And
Augusto B.\ Corr\^{e}a\\
University of Oxford\\
United Kingdom
\And
Felipe Meneguzzi\\
University of Aberdeen\\
United Kingdom
\And
Jendrik Seipp\\
Link{\"o}ping University\\
Sweden
}

\begin{document}
\maketitle
\makeatletter\global\@topnum\z@\makeatother
\input{abstract}

\input{introduction}
\input{background}
\input{proposed-method}
\input{heuristic-cases}
\input{experiments}
\input{related-work}
\input{conclusion}

\ificlrfinal
\section*{Acknowledgments}
This work was supported by the Wallenberg AI, Autonomous Systems and Software Program (WASP) funded by the Knut and Alice Wallenberg Foundation.
Andr\'{e} G.~Pereira acknowledges support from FAPERGS with projects 21/2551-0000741-9 and 25/2551-0002590-7.
This study was financed in part by the \textit{Coordena\c{c}\~ao de Aperfei\c{c}oamento de Pessoal de N\'{i}vel Superior -- Brasil} (CAPES) -- Finance Code~001.
\fi

\section*{AI Use Statement}
We used LLM coding agents to develop the heuristic programs and Lean proofs described in Section~\ref{sec:method}.
We also used AI assistants to check the literature, review and revise the manuscript and prepare the analysis and figures.

\section*{Ethics Statement}
This work uses public planning benchmarks and three new benchmark domains (Appendix~\ref{app:new-domains}), and involves no human subjects or personal data.
LeanPlan's guarantees refer to the PDDL model, so an inaccurate model can still yield plans that are unsafe or suboptimal for the real system it describes.
Developing each heuristic and proof requires LLM computation, but LeanPlan reuses every accepted heuristic on new tasks of its domain without further LLM calls.

\section*{Reproducibility Statement}
Appendix~\ref{app:proofs} specifies the proof interface and search conditions, and Appendix~\ref{app:heuristics} describes the heuristics for all thirteen domains.
Appendix~\ref{app:records} documents the run records and their analysis.
We will make the source code, benchmarks and run logs publicly available upon acceptance.

\bibliographystyle{iclr2027_conference}
\bibliography{./abbrv,./literatur,./biblio,./crossref-short}
\clearpage
\appendix
\input{appendix}
\end{document}

%% file: commands.tex
\renewcommand{\cite}[1]{\citep[][]{#1}}

\newcommand{\astar}{\ensuremath{\mathrm{A}^{*}}}

\newcommand{\hstar}{\ensuremath{h^{*}}}

\makeatletter
\newcommand{\mypar}{\@startsection{paragraph}{4}{\z@}{0.5ex plus 0.2ex minus .1ex}{-1em}{\normalsize\bf}}
\makeatother

%% file: abstract.tex
\begin{abstract}
Frontier large language models (LLMs) can generate \emph{heuristic functions} that guide search to achieve state-of-the-art performance in \emph{satisficing} planning, where any plan is acceptable.
However, these heuristics are not guaranteed to be \emph{admissible} and can lead to suboptimal plans.
We introduce LeanPlan, the first planning system that finds optimal plans with LLM-generated heuristics whose admissibility is machine-checked.
Given a domain description and training tasks, an agentic loop uses planner feedback to iteratively improve a reusable domain-specific heuristic, its admissibility proof and the required domain assumptions.
LeanPlan implements the heuristic, its proof and an efficient planner with machine-checked grounding and search in Lean\,4.
We evaluate LeanPlan on ten domains from the International Planning Competition and three new domains, using test tasks with up to 57 times as many objects as the training tasks.
With GPT-5.6 Sol in the agentic loop, we successfully generate heuristics and admissibility proofs for all these domains.
With the resulting heuristics, LeanPlan usually expands fewer states than the state-of-the-art Scorpion planner and solves more tasks overall.
\end{abstract}

%% file: introduction.tex
\section{Introduction}

\input{figure-pipeline}

Classical planning asks for a sequence of actions that transforms an initial state into a state that satisfies a goal~\cite{ghallab-et-al-2004}.
Planning tasks come in \emph{domains}, families of tasks with the same kinds of objects and actions, such as stacking blocks or moving a lift between floors.
Domains and tasks are usually described in the Planning Domain Definition Language (PDDL)~\cite{haslum-et-al-2019}.
Heuristic search is one of the most successful approaches to classical planning~\cite{bonet-geffner-aij2001,hoffmann-nebel-jair2001}.
A \emph{heuristic} estimates the cost of reaching a goal from a state, and the search uses these estimates to decide which states to explore next~\cite{pearl-1984}.
Domain-independent heuristics derive their estimates from the task description alone and therefore apply to any task.
Domain-specific heuristics exploit knowledge about a particular domain and can be much more informative~\cite{junghanns-schaeffer-aij2001}, but a human expert usually has to design and implement each of them.

Large language models (LLMs) can now automate the design of such heuristics.
\citet{correa-et-al-neurips2025} generate heuristic programs once per domain and select them on training tasks.
With greedy best-first search, their strongest heuristics solve more unseen tasks in total than strong domain-independent heuristics.
Other approaches repair candidate heuristics with counterexamples to a property checked on the training tasks~\cite{pereira-et-al-neurips2026}, generate heuristics for tasks given by successor generators and goal tests~\cite{tuisov-et-al-icaps2026} or evolve domain-independent heuristic programs~\cite{gestrin-seipp-icaps2026wslm4plan}.
LLMs can also generate generalized plans, i.e., programs that produce plans for new tasks of a domain~\cite{silver-et-al-aaai2024,stein-et-al-icaps2026,murray-et-al-icaps2026}.

However, these approaches focus on \emph{satisficing planning}, where any valid plan suffices.
In \emph{optimal planning}, we require a plan of minimum cost.
\astar{} search~\cite{hart-et-al-ieeessc1968} provides this guarantee when its heuristic is \emph{admissible}, meaning that it never overestimates the remaining plan cost.
Admissibility is hard to establish, because it must hold in every state of every task of the domain, including tasks much larger than the training tasks.
Neither testing on example tasks nor learning heuristic values from solved examples~\cite{chen-et-al-icaps2024} can establish this property.
\citet{futuhi-sturtevant-iclr2026} study near-admissible neural heuristics with statistical generalization bounds, but these heuristics can still overestimate, so \astar{} may return suboptimal plans.
\citet{phung-et-al-icaps2026wsripl} instead guarantee admissibility by construction: they evolve programs that choose task abstractions, from which pattern databases and cost partitioning derive admissible heuristics.

In this paper, we study a complementary approach: generating the heuristic program together with a proof of its admissibility.
We introduce \emph{LeanPlan}, the first planning system that finds optimal plans with LLM-generated heuristics whose admissibility is machine-checked.
LeanPlan implements each domain heuristic and its proof in Lean~4~\cite{moura-ullrich-cade2021}, a programming language and proof assistant, so that Lean's kernel checks the proof.
The heuristic can be any Lean program, for example one that combines domain-specific action counts, distance bounds and dead-end tests, as long as its admissibility can be proved.
Because the proof is about the code that the planner actually runs, there is no separate implementation that could deviate from the verified one.
Shared proofs, which apply to all domains, turn each admissibility proof into an optimality guarantee for the plans that the planner returns.
Unlike a statistical generalization bound, the proof covers every task of the domain that satisfies its checked assumptions, however large, and not only the training tasks.

Figure~\ref{fig:pipeline} shows how LeanPlan develops and uses a heuristic.
A deterministic generator first translates the PDDL domain into a Lean module, which fixes the vocabulary available to the heuristic and its proof.
Given this module, the domain's training tasks and a shared proof library, an LLM coding agent writes the heuristic, its admissibility proof and Boolean checks of the domain assumptions that the proof relies on.
The agent builds the code and runs the planner itself, uses failures as feedback and revises the artifacts until it judges that it cannot improve the heuristic further.
Whether the proof is correct is decided by Lean's kernel, a small trusted proof checker, and not by the agent's own report of success.
An accepted heuristic then serves \astar{} on new tasks of the domain, without further LLM calls or new proofs.

We evaluate LeanPlan on 900 test tasks from ten domains of the Learning Track of the 2023 International Planning Competition (IPC)~\cite{taitler-et-al-aimag2024} and on 270 test tasks from three new domains.
The new domains were not publicly available before our experiments, which rules out exposure through public LLM training data.
In IPC 2023, the hard test tasks have 2.5 to 57 times as many objects as the largest training task of their domain, compared with at most about twice as many in the new domains.
With GPT-5.6 Sol as the coding agent, LeanPlan obtains an admissible heuristic with a machine-checked proof for all thirteen domains.
With the generated heuristics, LeanPlan solves 281 IPC 2023 tasks and 67 new-domain tasks, 27.1\% and 31.4\% more than the state-of-the-art optimal planner Scorpion~\cite{seipp-ecai2024}, which solves 221 and 51 tasks.
A planner with machine-checked optimality guarantees can thus outperform a state-of-the-art, domain-independent optimal planner that offers no such guarantees.

%% file: figure-pipeline.tex
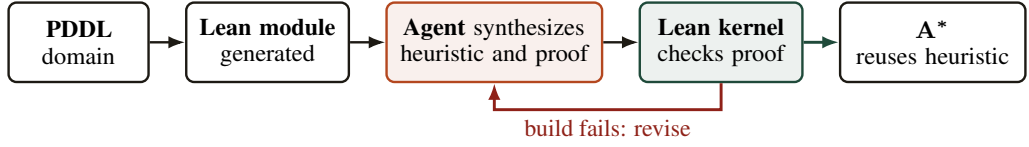
\begin{figure}[t]
\centering
\begin{tikzpicture}
  \node[stage, text width=1.5cm] (pddl) at (0,0) {\textbf{PDDL}\\domain};
  \node[stage, text width=1.8cm] (lean) at (2.5,0) {\textbf{Lean module}\\generated};
  \node[stageA, text width=2.5cm] (ag) at (5.5,0)
    {\textbf{Agent} synthesizes\\heuristic and proof};
  \node[stageG, text width=1.8cm] (k) at (8.5,0) {\textbf{Lean kernel}\\checks proof};
  \node[stage, text width=2.1cm] (as) at (11.3,0) {\textbf{\boldmath\astar{}}\\reuses heuristic};
  \draw[ar] (pddl) -- (lean);
  \draw[ar] (lean) -- (ag);
  \draw[ar] (ag) -- (k);
  \draw[arG] (k) -- (as);
  \draw[arR] (k.south) -- ++(0,-0.35) -| (ag.south);
  \node[font=\footnotesize, color=cRed, anchor=north] at (7.0,-0.9) {build fails: revise};
\end{tikzpicture}
\caption{Proof-guided heuristic development and reuse in LeanPlan. A deterministic generator translates the domain into a Lean module, and an agent revises a heuristic and its proof using feedback from Lean. %
At planning time, task checks enforce the proof assumptions.}
\label{fig:pipeline}
\end{figure}

%% file: background.tex
\section{Background}
\label{sec:background}

\mypar{Classical Planning.}
We consider classical planning, where actions are deterministic and states are fully observable~\cite{ghallab-et-al-2004}.
The Planning Domain Definition Language (PDDL)~\cite{haslum-et-al-2019} separates a reusable \emph{domain} from an individual \emph{task}, usually in two files.
The domain declares object types, predicates over objects and \emph{action schemas}, i.e., actions with object parameters.
The task supplies the objects, the initial state and the goal.
The same domain can therefore describe many tasks with different objects, initial states and goals.
\emph{Grounding} instantiates the predicates and action schemas with the task's objects to obtain \emph{atoms} and \emph{ground actions}.
Together with the task's initial state and goal, they form a finite \emph{ground task}, which the planner solves.

Ground tasks use the STRIPS representation~\cite{fikes-nilsson-aij1971}.
It represents a \emph{state} by the set of atoms that are true in it.
Each ground action~$a$ has a cost~$c(a)$ and three sets of atoms: its preconditions, add effects and delete effects.
It is applicable in a state~$s$ if all its preconditions are true in~$s$.
Applying it first removes its delete effects and then adds its add effects, leaving all other atoms unchanged.
We write $s\xrightarrow{a}s'$ for the resulting transition from~$s$ to its successor state~$s'$.
The goal is a set of atoms, and a \emph{goal state} is a state in which all of them are true.
A state is \emph{reachable} if a sequence of transitions leads to it from the initial state.

A \emph{plan} for a state~$s$ is a sequence of actions that can be applied one after the other, starting in~$s$, and that leads to a goal state.
Its cost is the sum of its action costs, and an \emph{optimal} plan for~$s$ has minimum cost among all plans for~$s$.
We write $\hstar(s)$ for the cost of an optimal plan for~$s$.
If $s$ has no plan, it is a \emph{dead end} and $\hstar(s)=\infty$.
Optimal planning asks for an optimal plan for the initial state.
In our experiments, every action has cost one, so an optimal plan also has minimum length.

\mypar{Heuristic Search.}
A \emph{heuristic}~$h$ estimates $\hstar$: it maps each state~$s$ to a nonnegative number or $\infty$.
It is \emph{admissible} if $h(s)\leq\hstar(s)$ for every state~$s$, so an admissible heuristic returns $\infty$ only for dead ends.
\astar{} search~\cite{hart-et-al-ieeessc1968} starts from the initial state and repeatedly \emph{expands} a state, i.e., generates its successors.
It always expands a state that minimizes the cost of the cheapest known path to it plus its heuristic value, and it prunes states whose heuristic value is $\infty$.
With an admissible heuristic, \astar{} returns an optimal plan for the initial state whenever one exists~\cite{hart-et-al-ieeessc1968}.

Admissibility is a global property, since $\hstar(s)$ depends on all plans for~$s$.
We therefore prove two local properties, each concerning a single state or transition, that together imply admissibility.
A heuristic is \emph{goal-aware} if $h(s)=0$ for every goal state~$s$, and \emph{consistent} if $h(s)\leq c(a)+h(s')$ for every transition $s\xrightarrow{a}s'$.
Along any plan, consistency lets the heuristic value decrease by at most the cost of each action, and goal awareness makes the final value zero.
The value at the start of the plan is therefore at most the plan's cost, so a goal-aware and consistent heuristic is admissible.
LeanPlan's domain proofs establish both properties only for reachable states.
This suffices because every state on a plan for a reachable state is itself reachable, so the argument above yields admissibility on all reachable states.

\mypar{Lean Programs and Proofs.}
Lean~4 combines a functional programming language with an interactive theorem prover~\cite{moura-ullrich-cade2021}.
Its type system expresses both executable functions and logical statements about them.
For example, a function from natural numbers to natural numbers has type \texttt{Nat -> Nat}, while the proposition \texttt{forall n : Nat, n + 0 = n} has type \texttt{Prop}, the type of propositions.
Because programs and propositions share one language, a theorem can refer directly to the definition of a total function (one that terminates on every input), including its arithmetic, case distinctions and recursive calls.
A proof of a proposition is a \emph{proof term} whose type is that proposition.
Lean's \emph{kernel}, a small checker for Lean's core logic, verifies that the term has the claimed type, using only the declared definitions and axioms.

Proof terms are usually constructed with \emph{tactics} instead of being written directly.
A \emph{proof goal} consists of a proposition to establish and the local assumptions available for proving it.
A tactic transforms a proof goal into zero or more simpler goals while building the corresponding proof term, for example by unfolding definitions, applying existing lemmas, splitting cases, performing induction or solving arithmetic subgoals.
The proof is complete when no goals remain.
The kernel then checks the resulting term independently of how it was constructed, so trusting a proof requires trusting only the kernel and the axioms.

A theorem holds for all values of its parameters that satisfy its assumptions.
A theorem about all natural numbers thus covers infinitely many values, not only those in a finite test set.
Likewise, LeanPlan's domain theorems cover every task that satisfies their assumptions (Section~\ref{sec:method}), not only the training tasks.
Theorems are separate declarations that the compiled planner never executes, so the proofs add no runtime cost.
Figure~\ref{fig:boarding} in Section~\ref{sec:cases} shows a small example of a function, a theorem about it and a tactic proof.

%% file: proposed-method.tex
\section{LeanPlan}
\label{sec:method}

LeanPlan develops a heuristic and its admissibility proof once per domain and reuses both for every new task of that domain (Figure~\ref{fig:pipeline}).
It accepts the unit-cost STRIPS fragment of PDDL described in Section~\ref{sec:background}, with optional typing, type hierarchies and domain constants.
Appendix~\ref{app:pddl} shows a Miconic domain and task in this fragment.
For each domain, a deterministic generator translates the PDDL action schemas into a Lean module, which fixes the vocabulary available to the heuristic and its proof.

\mypar{Domain Assumptions.}
A heuristic may rely on properties of a domain's tasks that PDDL does not enforce.
For example, a Blocksworld heuristic may assume that at most one block is directly on top of each block.
This property holds in every Blocksworld task of interest, but a syntactically valid task can violate it.
Following \citet{abdulaziz-et-al-icaps2022wshsdip}, we make such properties explicit as \emph{domain assumptions}, which are Boolean checks on the task's objects, initial state and goal.
The agent that develops the heuristic is free to choose the domain assumptions that its proof needs, as long as all training tasks satisfy them.
We call the collection of a heuristic's domain assumptions its \emph{certificate}.
The certificate also checks that the input domain has exactly the action schemas of the generated Lean module, since the proof reasons about these schemas.
LeanPlan evaluates the certificate before search and refuses to use the heuristic on a task that fails any of the certificate's checks.\footnote{These checks revealed a bug in the IPC 2023 Learning Track training set: the initial state of Floortile task \texttt{p08.pddl} contains both \texttt{(robot-at robot2 tile\_2\_1)} and \texttt{(clear tile\_2\_1)}, so another robot can move onto the occupied tile.
We reported the bug to the competition organizers, and it has since been fixed in the benchmark repository.}

\mypar{Proof Obligations.}
Let $d$ and $p$ be a domain and task that pass LeanPlan's validators, let $T$ be their ground task and let $h$ be the compiled heuristic that the search evaluates.
Let $R$ be the set of states reachable in~$T$.
Assuming that the certificate holds for~$p$, each domain theorem proves
\begin{align}
h(s)=0 &&\text{for every goal state }s\in R,\label{eq:goal-contract}\\
h(s)\leq c(a)+h(s') &&\text{for every transition }s\xrightarrow{a}s'\text{ with }s\in R.\label{eq:step-contract}
\end{align}
A shared theorem then derives admissibility on~$R$ with the argument from Section~\ref{sec:background}.
Appendix~\ref{app:proofs} gives the Lean interface and its connections to grounding and search.

\mypar{Generating Heuristics and Proofs.}
An LLM coding agent receives the generated Lean module, the domain's training tasks, the heuristic interface and the shared proof library.
It develops a heuristic and proves that the heuristic satisfies Equations~\ref{eq:goal-contract} and~\ref{eq:step-contract} whenever its certificate holds.
The agent builds the code and proofs and runs the planner on the training tasks, using failures as feedback for revision.
The prompt specifies:
\begin{itemize}
\item the objective: develop a domain-specific admissible heuristic with the best training performance the agent can achieve, continuing to refine it after beating a baseline planner;
\item the interface and context of the planner;
\item the correctness requirements: prove goal awareness and consistency for the compiled heuristic on reachable states under its certificate, and ensure that the certificate accepts every training task;
\item the baseline: measured Scorpion results, its search configuration with $h^2$-preprocessing, and the same per-task wall-clock limit and pinned CPU core for both planners;
\item a list of useful commands;
\item the prohibited shortcuts (e.g., \texttt{sorry}, \texttt{admit}, new axioms; see next paragraph);
\item the stopping rule: stop when the agent judges that it cannot improve the heuristic further, leaving its best candidate with a working proof and passing certificates.
\end{itemize}
Appendix~\ref{app:prompt} reproduces the full template from the agentic-loop implementation.

\mypar{Accepting a Candidate.}
Since the agent's own report of success is not evidence, an automated controller re-checks the files the agent leaves behind.
It rejects the candidate if these files declare a new axiom or use one of the Lean constructs \texttt{sorry}, \texttt{admit} and \texttt{native\_decide}, which skip a proof or trust unverified code.
It then rebuilds the generated Lean module, the proofs and the planner, and it checks that the domain theorems depend only on Lean's standard axioms and that the certificate accepts every training task.
Finally, it runs the planner on the training tasks.
The controller accepts the heuristic only if all checks pass and \astar{} with it solves at least as many training tasks as Scorpion~\cite{seipp-ecai2024}, while expanding strictly fewer states in total on the tasks both solve.
This comparison requires a nonempty shared set with complete expansion counts, even when LeanPlan solves more tasks.

\mypar{Grounding.}  To solve a new task, LeanPlan parses and validates it,
grounds it and, if the task passes the certificate, runs \astar{} with the
heuristic.  Grounding is similar to the algorithm by \citet{helmert-aij2009},
which is used in popular planners such as Fast Downward~\cite{helmert-jair2006}
and Scorpion~\cite{seipp-ecai2024}, the state-of-the-art optimal planner that extends Fast Downward. Grounding
instantiates only actions whose static preconditions, which no action can
change, hold in the initial state.  LeanPlan then prunes actions that $h^2$
mutexes rule out~\cite{alcazar-torralba-icaps2015} and removes effects and
actions that cannot influence the goal.

\mypar{Search.}
LeanPlan's \astar{} is a standard implementation with duplicate detection, reopening and goal tests at expansion time.
Like Scorpion, it breaks ties between states with the same sum of path cost and heuristic value in favor of lower heuristic values~\cite{asai-fukunaga-jair2017}.
Unlike Scorpion, it then prefers the most recently added state, which tends to be more efficient~\cite{asai-fukunaga-aaai2016}.

\mypar{Search Guarantees.}
Shared theorems extend the domain guarantees from the heuristic to the whole search.
Every plan that \astar{} returns is valid for the ground task, and with an admissible, goal-aware heuristic, it is optimal whenever it costs less than a large bound~$B$, far above the cost of any plan in our experiments.
Since heuristics in Lean return natural numbers, LeanPlan represents~$\infty$ by~$B$ and prunes every state whose heuristic value reaches it.
The proofs therefore certify optimal plans but not dead ends or unsolvable tasks, and Appendix~\ref{app:proofs} states the theorems and explains this limitation.

%% file: heuristic-cases.tex
\section{Domain Heuristics}
\label{sec:cases}

The generated heuristics use domain structure to count actions that every solution must take.
We illustrate this with three domains that show action counting, dead-end detection and distance-based relaxations, and Appendix~\ref{app:heuristics} summarizes the remaining heuristics and representation details.

\mypar{Miconic.}
In Miconic, a lift transports passengers to their destination floors.
The heuristic considers only passengers whose \texttt{served} goal is unmet, and it lets $W(s)$ contain those still waiting and $R(s)$ those aboard.
A waiting passenger must board the lift and leave it at the destination with the domain's \texttt{depart} action, while a passenger aboard needs only to leave.
Let~$F(s)$ be the set of their destination floors and the origin floors of $W(s)$, excluding the current lift floor, so that a floor needed by several passengers counts once.
The resulting heuristic is $2|W(s)|+|R(s)|+|F(s)|$.
A move reaches at most one required floor, and boarding or leaving reduces the passenger term $2|W(s)|+|R(s)|$ by exactly one without reducing $F(s)$, since it happens at the current floor, which $F(s)$ excludes.
Under the certified location and goal assumptions, each action therefore reduces the heuristic by at most one, so it is consistent, and all three sets are empty at a goal, so it is goal-aware.

Figure~\ref{fig:boarding} isolates the arithmetic step for boarding as a standalone Lean example.
\begin{figure}[t]
\begin{lstlisting}[language=Lean]
def passengerBound (w r : Nat) : Nat := 2 * w + r

theorem boarding_step (w r : Nat) :
    passengerBound (w + 1) r =
      1 + passengerBound w (r + 1) := by
  unfold passengerBound
  omega
\end{lstlisting}
\caption{A Lean example of a function and a theorem about it with a tactic proof. For a lift with $w$ waiting and $r$ riding passengers, \texttt{passengerBound} is a lower bound on the number of remaining boarding and leaving actions, and \texttt{boarding\_step} shows that boarding reduces this bound by exactly one. The \texttt{def} declaration gives the function's parameters, return type and body, and the \texttt{theorem} declaration states an equality about the function for all natural numbers \texttt{w} and \texttt{r}. The block after \texttt{by} constructs its proof: \texttt{unfold} replaces the function by its definition, and \texttt{omega} closes the resulting linear arithmetic goal.}
\label{fig:boarding}
\end{figure}
This step alone does not establish consistency, since the domain proof must also justify the count updates from the state transition and show that boarding does not reduce the floor term under the domain invariants.

\mypar{Spanner.}
A worker walks along a directed path to a gate with loose nuts that need to be tightened.
On the path, the worker can collect spanners, each of which can tighten only one nut.
The heuristic combines action counting with dead-end detection.
Let $L(s)$ be the number of goal nuts that are still loose and $C(s)$ the number of distinct usable spanners that the worker carries.
Let $D(s)$ be the directed distance from the worker to the gate.
Tightening the nuts requires $L(s)$ actions and collecting the missing spanners at least $\max(0,L(s)-C(s))$ pickups, and adding the movement bound gives $L(s)+\max(0,L(s)-C(s))+D(s)$.
The heuristic is zero when all goal nuts are tight.
It returns the pruning threshold~$B$ if $C(s)$ plus the number of distinct usable spanners still reachable on the ground is less than $L(s)$, which detects dead ends with too few usable spanners.

\mypar{Sokoban.}
In this Sokoban formulation, each goal requires a specific box on a designated square.
The heuristic is the maximum of two lower bounds that both ignore the other boxes when moving a box.
Both use the push graph, which connects two squares if a box can be pushed from one to the other while the robot stands behind the box.
The first bound adds up pushes and walking.
For each box $b$ with an unmet goal, let $P_b(s)$ count the pushes on a shortest path to its square in the push graph.
Since each push moves only one box, different boxes need different pushes, so the sum of these counts is a lower bound on the number of pushes.
Before its first push, the robot must also walk to a square from which it can push some box, and $A(s)$ is the length of the shortest such walk.
Since the first push may move a box that is already at its goal, $A(s)$ also considers such boxes.
The second bound considers each box on its own, with $J_b(s)$ counting the robot moves and pushes on a shortest path that brings $b$ to its square when only $b$ and the robot remain.
The two bounds may count the same actions, so the heuristic takes their maximum instead of their sum:
\[
h(s)=\max\left\{\sum_{b\text{ with unmet goal}} P_b(s)+A(s),\quad
                       \max_{b\text{ with unmet goal}} J_b(s)\right\}
\]
When no box goal is unmet, the heuristic sets all terms, including $A(s)$, to zero, so it is goal-aware.
LeanPlan computes the distance tables with breadth-first search when it compiles the heuristic.
Instead of proving this search correct, it checks that each table has the properties the proof relies on, such as distances that decrease by at most one per move.

%% file: experiments.tex
\section{Experimental Results}
\label{sec:experiments}

We use the ten domains from the IPC~2023 Learning Track~\cite{taitler-et-al-aimag2024} and three new domains, Orrery, Cisterns and Qubit Routing, described in Appendix~\ref{app:new-domains}.
The new domains and their tasks were not publicly available before the experiments, so the LLM cannot have seen them in public training data.
Each domain has 90 test tasks, for 900 IPC 2023 tasks and 270 tasks in the new domains.
The heuristics are developed on each domain's training tasks, 89 to 99 per IPC 2023 domain and 20 per new domain.
We run GPT-5.6 Sol at its highest reasoning effort as the coding agent in the Codex harness to generate heuristics and prove their admissibility.
The agent runs in a sandbox that prevents access to the test tasks and restricts file modifications to the designated workspace.
\jendrik{Shall we mention Codex or not?}

Across our runs, the agentic loop takes an average of 69.8\,minutes per domain (standard deviation 16.5\,minutes), excluding time spent evaluating the baseline planner.
It uses an average of 33\,724\,876 input tokens, including cached input, and 103\,196 output tokens per domain.
About 98.6\% of the input tokens are cached.
The agent tries 10.3 heuristic variants per domain on average and at most 13.
At API prices of US\$4, US\$0.40 and US\$20 per million uncached input, cached
input and output tokens, the average cost is US\$17.30 per domain (standard
deviation US\$4.62).  Given the time and monetary cost of
repeated runs, we prioritize evaluating a broad set of thirteen domains over
running the agentic loop multiple times per domain.

We compare five planner configurations.
Two use LeanPlan: \emph{blind LeanPlan} with the blind heuristic (0 at goal states and 1 elsewhere) and \emph{LeanPlan} with the generated heuristics.
The other three use the state-of-the-art optimal planner Scorpion~\cite{seipp-ecai2024}: \emph{blind Scorpion} with the blind heuristic, \emph{LM-cut} with the domain-independent LM-cut heuristic~\cite{helmert-domshlak-icaps2009} and \emph{SCP} with Scorpion's recommended configuration.
SCP combines abstraction heuristics via saturated cost partitioning~\cite{seipp-et-al-jair2020}, and we adjust its component budgets to our resource limit.\footnote{The
\href{https://github.com/jendrikseipp/scorpion}{recommended Scorpion configuration}
allocates up to 100\,seconds to systematic pattern collection construction,
with 10\,seconds per restart, and 1\,000\,seconds to online cost partitioning.
We keep the per-restart limit and use 10 and 150\,seconds for the other two
budgets, which gave the best coverage in earlier parameter sweeps on
the test suite under our 300-second overall limit.
During heuristic development, the agentic loop gives both planners
30\,seconds per training task and lowers all three budgets to 1, 1 and
15\,seconds (Appendix~\ref{app:prompt}).
}
All five read the same PDDL inputs and run \astar{} after pruning actions that $h^2$ mutexes rule out~\cite{alcazar-torralba-icaps2015}.
We run each configuration once per task with Downward Lab~\cite{seipp-et-al-zenodo2017}, limiting every run to 300\,CPU seconds and 8\,GiB of memory.
We report \emph{coverage}, the number of tasks solved, as well as planner time and the number of states expanded during search.
The plan validator VAL~\cite{howey-long-icaps2003wscompetition} accepts every returned plan, and all planners that solve a task find plans of the same cost.

\input{tables/coverage}

\subsection{Blind Search}
\label{sec:blind}

Since the blind heuristic gives no guidance, comparing blind LeanPlan with blind Scorpion measures how efficiently the two planners implement grounding and search.
Blind LeanPlan solves nearly as many tasks as blind Scorpion (Table~\ref{tab:coverage}).
The two planners solve the same number of tasks in all domains except Rovers and Cisterns, and every task solved by blind LeanPlan is also solved by blind Scorpion.
On the IPC 2023 tasks that both solve, they expand almost the same number of states in total (differences are due to tie-breaking), but LeanPlan's search takes about 1.5 times as long.
The remaining gap therefore stems from slower state expansion rather than from different search behavior.
Slower expansion also explains why blind LeanPlan solves fewer tasks within one second (106 compared to 116), although it nearly catches up within ten seconds (138 compared to 139).
Overall, proof-checked grounding and search cost little coverage compared with Scorpion, whose Fast Downward code base~\cite{helmert-jair2006} has been optimized for many years.
\jendrik{"the grounding time that LeanPlan's timer includes but Scorpion's does not". Include grounding time in Scorpion's time.}

\subsection{\texorpdfstring{\astar{}}{A*} Search with Heuristics}
\label{sec:informed}

With the generated heuristics, LeanPlan solves the most IPC 2023 tasks (Table~\ref{tab:coverage}).
Its largest gain is in Spanner, which we attribute to the dead-end test of its heuristic, but even without Spanner, it solves more tasks than SCP (209 compared to 191).
Still, SCP solves more tasks than LeanPlan in four IPC 2023 domains, so the generated heuristics do not dominate domain-independent ones everywhere.
LeanPlan's advantage is largest on tasks beyond the training sizes.
Of the 595 IPC 2023 test tasks with more objects than every training task of their domain, LeanPlan solves 53, compared to 13 for SCP, 7 for LM-cut and 0 for blind search.
The generated heuristics thus remain effective far outside the training distribution.

\input{figure-performance}

The left panel of Figure~\ref{fig:performance} shows how quickly each configuration solves tasks.
LeanPlan and LM-cut need little precomputation, so their curves rise from the start.
LM-cut's curve flattens early, however, because its heuristic is expensive to evaluate: on the tasks that both solve, LM-cut's search takes five times as long as LeanPlan's.
SCP first spends up to ten seconds constructing its pattern collection, so its curve rises only afterwards.
Within ten seconds, LeanPlan solves more tasks than SCP within the full time limit.

The right panel of Figure~\ref{fig:performance} compares the expansions of LeanPlan and SCP on the 211 IPC 2023 tasks that both solve.
LeanPlan expands fewer states on 152 of these tasks and more on 55, and the balance depends on the domain.
In Spanner, where LeanPlan's heuristic prunes dead ends, LeanPlan expands fewer states on every task.
In Sokoban, SCP expands fewer states on most tasks, presumably because its abstractions capture some interactions between boxes, which LeanPlan's relaxations ignore.
Since the planners also differ in other components, such as tie-breaking, these counts compare complete configurations rather than heuristics alone.
Compared with blind LeanPlan, the generated heuristics reduce the number of expansions on every task that both solve, by more than two orders of magnitude in total.

In the new domains, blind search solves only a few Cisterns tasks, and LM-cut solves few tasks in any domain.
LeanPlan solves more tasks than SCP overall, especially in Qubit Routing, but one task fewer in Cisterns.
There, the certificate refuses three test tasks, the only refusals among all 1\,170 test tasks, so the proof does not cover them and LeanPlan counts them as unsolved.
SCP solves one of them, which accounts for its lead.
No configuration solves the other two.
Since the certificate must accept all training tasks, a more varied training set could have led the controller to reject this heuristic and the agent to find one that also covers such tasks.

%% file: tables/coverage.tex
\begin{table}[t]
\centering
\small
\caption{Tasks solved, with 90 per domain and separate sums for IPC 2023 and the new domains. Total combines both suites. The left columns use blind \astar{} search and the right columns use \astar{} search with heuristics. SCP is Scorpion's recommended configuration, which uses saturated cost partitioning \cite{seipp-et-al-jair2020}. All configurations prune actions with $h^2$ mutexes \cite{alcazar-torralba-icaps2015}. Bold values and shading mark the highest coverage in each row, including ties.}
\label{tab:coverage}
\setlength{\tabcolsep}{4.5pt}
\begin{tabular}{@{}clrrcrrr}
\toprule
& & \multicolumn{2}{c}{Blind \astar{}} & & \multicolumn{3}{c}{\astar{} with heuristics} \\
\cmidrule(lr){3-4}\cmidrule(lr){6-8}
&  &  &  & & Scorpion & Scorpion &  \\
& Domain & Scorpion & LeanPlan & & LM-cut & SCP & LeanPlan \\
\midrule
\multirow{10}{*}{\rotatebox[origin=c]{90}{IPC 2023}} & Blocksworld & 6 & 6 &  & 11 & 11 & \cellcolor{red!15}\textbf{20} \\
 & Childsnack & 9 & 9 &  & 9 & 9 & \cellcolor{red!15}\textbf{16} \\
 & Ferry & 10 & 10 &  & 18 & 19 & \cellcolor{red!15}\textbf{29} \\
 & Floortile & 10 & 10 &  & \cellcolor{red!15}\textbf{20} & \cellcolor{red!15}\textbf{20} & \cellcolor{red!15}\textbf{20} \\
 & Miconic & 30 & 30 &  & 36 & \cellcolor{red!15}\textbf{40} & 37 \\
 & Rovers & 15 & 12 &  & 16 & 17 & \cellcolor{red!15}\textbf{18} \\
 & Satellite & 12 & 12 &  & 20 & \cellcolor{red!15}\textbf{27} & 25 \\
 & Sokoban & 27 & 27 &  & 29 & \cellcolor{red!15}\textbf{31} & 30 \\
 & Spanner & 30 & 30 &  & 30 & 30 & \cellcolor{red!15}\textbf{72} \\
 & Transport & 8 & 8 &  & 9 & \cellcolor{red!15}\textbf{17} & 14 \\
\cmidrule(l){2-8}
 & Sum & 157 & 154 &  & 198 & 221 & \cellcolor{red!15}\textbf{281} \\
\midrule
\multirow{3}{*}{\rotatebox[origin=c]{90}{New}} & Orrery & 0 & 0 &  & 1 & 18 & \cellcolor{red!15}\textbf{20} \\
 & Cisterns & 9 & 8 &  & 10 & \cellcolor{red!15}\textbf{26} & 25 \\
 & Qubit Routing & 0 & 0 &  & 0 & 7 & \cellcolor{red!15}\textbf{22} \\
\cmidrule(l){2-8}
 & Sum & 9 & 8 &  & 11 & 51 & \cellcolor{red!15}\textbf{67} \\
\midrule
 & Total & 166 & 162 &  & 209 & 272 & \cellcolor{red!15}\textbf{348} \\
\bottomrule
\end{tabular}
\end{table}

%% file: figure-performance.tex
\begin{figure}[t]
\centering
\includegraphics[width=\linewidth]{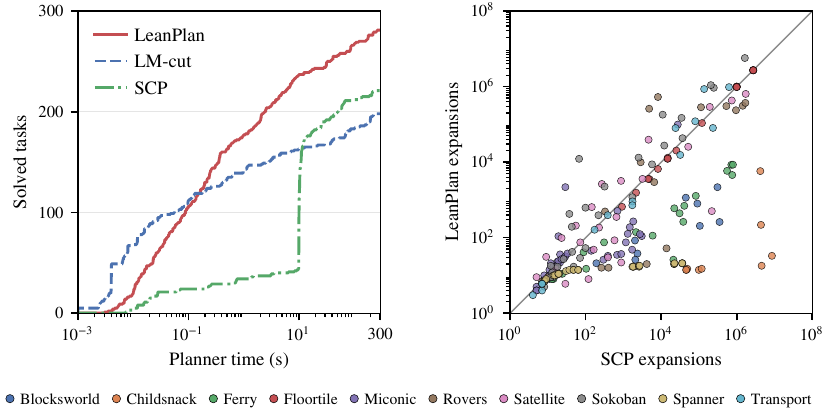}
\caption{\astar{} search with heuristics on IPC 2023.
Left: number of tasks solved within each reported per-task time threshold, with a logarithmic horizontal axis.
Right: expansions of LeanPlan and SCP on the 211 tasks that both solve, with logarithmic axes.
Each point represents one task, points below the diagonal favor LeanPlan and colors identify domains.}
\label{fig:performance}
\end{figure}

%% file: related-work.tex
\section{Related Work}
\label{sec:related}

\mypar{LLMs for Planning.}
Early studies found that LLMs cannot reliably generate valid plans, even with chain-of-thought prompting~\cite{valmeekam-et-al-neurips2023,stechly-et-al-neurips2024}.
Recent frontier models are competitive with classical planners on newly generated tasks from the IPC domains~\cite{correa-et-al-arxiv2025}, but their plans still require external validation.
Many approaches therefore use LLMs to produce artifacts that symbolic systems execute or check~\cite{kambhampati-et-al-icml2024}.
These artifacts include PDDL models~\cite{guan-et-al-neurips2023,tantakoun-et-al-acl2025}, successor functions and goal tests~\cite{katz-et-al-neurips2024} and the generalized plans and heuristic functions discussed in the introduction.

\mypar{LLM-Guided Program Search.}
Outside of planning, LLM-guided search finds programs that score well on a fixed set of instances~\cite{romera-paredes-et-al-nature2024,liu-et-al-icml2024,ye-et-al-neurips2024,novikov-et-al-arxiv2025}.
Some of these systems verify individual outputs, such as a cap set or a matrix multiplication algorithm, but none proves that a program has a property on instances outside this set.
Counterexample-guided inductive synthesis instead checks each proposed program against a formal specification~\cite{solar-lezama-et-al-asplos2006,abate-cav2018}.
LeanPlan combines both ideas: search on training tasks measures how informative a heuristic is, while the Lean kernel checks its admissibility.

\mypar{Learned Heuristics and Admissibility.}
Machine learning has long been used to obtain heuristics for search~\cite{arfaee-et-al-aij2011,agostinelli-et-al-nmi2019,shen-et-al-icaps2020,ferber-et-al-ecai2020,chen-et-al-aaai2024,chen-et-al-icaps2024}.
Such heuristics are generally inadmissible, and methods that aim for heuristics that are admissible with high probability~\cite{ernandes-gori-ecai2004,marom-rosman-icaps2020} or come with statistical generalization bounds~\cite{futuhi-sturtevant-iclr2026} do not guarantee admissibility on unseen states.
Domain-independent heuristics such as LM-cut~\cite{helmert-domshlak-icaps2009}, pattern databases~\cite{edelkamp-ecp2001,haslum-et-al-aaai2007}, merge-and-shrink~\cite{helmert-et-al-jacm2014} and saturated cost partitioning~\cite{seipp-et-al-jair2020} are admissible by construction, and \citet{phung-et-al-icaps2026wsripl} use LLMs to generate pattern generators within this family.

\mypar{Verification and Certificates in Planning.}
Plan validators such as VAL~\cite{howey-long-icaps2003wscompetition} check plans, \citet{abdulaziz-lammich-ictai2018} formally verify such a validator in Isabelle/HOL~\cite{nipkow-et-al-2002} and \citet{abdulaziz-kurz-aaai2023} formally verify a SAT-based planner.
Certifying algorithms~\cite{mcconnell-et-al-csr2011} return a certificate that an independent checker can verify, and certifying planners provide such certificates for unsolvability~\cite{eriksson-et-al-icaps2017,eriksson-et-al-icaps2018} and optimality~\cite{mugdan-et-al-icaps2023,dold-et-al-icaps2025}.
\citet{dold-et-al-icaps2025} extend \astar{} to emit pseudo-Boolean certificates of optimality during the search, which the independent proof checker VeriPB then validates.
These certificates are generated and checked for each task, whereas LeanPlan proves admissibility once per domain.
What we call a certificate (Section~\ref{sec:method}) is therefore not a per-task proof, but a fixed set of Boolean checks that establishes the premise of the domain theorems for each task.
Closest to our work, \citet{abdulaziz-et-al-icaps2022wshsdip} use Isabelle/HOL to prove for Miconic that a generalized potential heuristic~\cite{frances-et-al-ijcai2019} is descending on all tasks.
\citet{behnke-et-al-icaps2026wshsdip} formalize \astar{} and standard planning heuristics in Lean, using LLM-generated proofs for several components.
LeanPlan instead targets admissibility of generated domain-specific heuristics, and LLM coding agents construct their proofs without human interaction.

\mypar{Theorem Proving and Verified Code Generation.}
LLMs increasingly construct proofs in interactive theorem provers such as Lean and Isabelle~\cite{jiang-et-al-iclr2023,yang-et-al-neurips2023,xin-et-al-arxiv2024,ren-et-al-arxiv2025,hubert-et-al-nature2026}, and they can repair failed proofs using the prover's error messages~\cite{first-et-al-fse2023}.
Verifiable code generation asks for programs together with proofs of their correctness~\cite{aggarwal-et-al-icml2025,ye-et-al-iclr2026}.
For example, \citet{li-et-al-arxiv2026p3} first generate a joint plan for the program and its proof, then elaborate both in Lean.
LeanPlan specializes this idea to heuristic search, where every heuristic has the same proof obligations and where correctness alone is easy to achieve, since the heuristic that always returns zero is admissible.
The agent must therefore also make the heuristic informative, which we measure by search on training tasks.

%% file: conclusion.tex
\section{Conclusions}

We introduced LeanPlan, a planner in which LLM coding agents develop domain-specific heuristics together with machine-checked proofs of their admissibility.
For all domains considered, the agents produced such heuristics and proofs, so LLMs can help optimal planning without giving up the optimality guarantee.
Requiring proofs did not force the agents to settle for weak heuristics: the proved heuristics exploit domain structure, and with them LeanPlan usually expands fewer states than the state-of-the-art optimal planner Scorpion and solves more tasks overall.

\input{limitations}

%% file: limitations.tex
\mypar{Limitations.}
LeanPlan supports only the unit-cost STRIPS fragment of PDDL, and our evaluation uses thirteen domains and a single LLM and coding-agent harness.
The proofs guarantee admissibility only on tasks that satisfy the checked domain assumptions, and these assumptions can exclude test tasks, as in Cisterns.
They do not certify unsolvability.
The guarantees also rely on parts that the proofs do not check: Lean's kernel, compiler and runtime, LeanPlan's PDDL parser and the Lean definitions that formalize PDDL semantics.
Extending the approach to action costs and richer PDDL fragments and evaluating other LLMs, including open-weight ones, remain open directions.

%% file: appendix.tex
\section{PDDL Input Example}
\label{app:pddl}

Figure~\ref{fig:pddl} shows a Miconic domain excerpt and a complete task.
The \texttt{board} action puts a passenger aboard when the lift is at their origin floor.
\input{figure-pddl}

\section{Proof Interface and Search Conditions}
\label{app:proofs}

The shared interface in \texttt{Proofs/Heuristic.lean} expresses the contract in Equations~\ref{eq:goal-contract} and~\ref{eq:step-contract} as follows:
\begin{lstlisting}[language=Lean,breaklines=true]
structure ImprovedProof (d : Pddl.Domain) (p : Pddl.Problem)
    (rel : Bool) (check : Bool) (h : Heuristic) : Prop where
  goalAware : check = true ->
    (ground d p rel).GoalAwareOn (Reachable (ground d p rel)) h.eval
  consistent : check = true ->
    (ground d p rel).ConsistentOn (Reachable (ground d p rel)) h.eval
\end{lstlisting}
The listing replaces Lean's implication glyph with its ASCII notation.
The declaration \texttt{structure ... : Prop} defines a proposition with two proof fields, \texttt{goalAware} and \texttt{consistent}.
Constructing a proof of \texttt{ImprovedProof} requires supplying a proof of each field.
In each field, \texttt{check = true ->} introduces an assumption, and \texttt{h.eval} names the executable function whose property must be proved.
The Boolean \texttt{check} is therefore an input to the proposition, not a proof term.
At runtime, evaluating it establishes which tasks satisfy the theorem's premise without reconstructing the proof.
The flag \texttt{rel} selects whether grounding uses relevance pruning, and the domain theorems cover both values.
Each of the thirteen domain files exports the theorem \texttt{improved\_proof\_of\_certificate}, whose premise \texttt{Validated d p} requires both the domain and task validators to succeed.
The theorem instantiates \texttt{check} with the domain certificate and \texttt{h} with the runtime evaluator: \texttt{improved (ground d p rel)} for IPC 2023 and \texttt{Agent.heuristic (ground d p rel)} for the new domains.
The theorem \texttt{ImprovedProof.admissible} derives admissibility from these two fields and closure of reachability under transitions.
The theorem covers every accepted task and every state reachable in that task.

\mypar{From Domain Invariants to Executable Values.}
The proof establishes the initial invariants from the certificate and proves that the action schemas preserve them.
\texttt{reachable\_abstracts\_inv} in \texttt{Proofs/Reachable.lean} transfers these invariants to reachable packed states using the shared grounding lemmas.
\texttt{ComputesOn} in \texttt{Proofs/LiftedHeuristic.lean} equates a lifted heuristic with its compiled runtime value on the invariant states.
The domain proofs combine shared correspondence lemmas with domain-specific arguments to prove the contract for the compiled evaluator.
The shared grounding proofs account for atom numbering, static preconditions and relevance pruning.
Thus the proof applies to the function that search evaluates.

\mypar{Grounding Proof Boundary.}
The theorem \texttt{ground\_plan\_sound} transfers a plan of the packed-state task to the atom-level operators produced by \texttt{groundedOps}, preserving goal satisfaction and cost.
Separate lemmas establish schema-instantiation properties and plan preservation under relevance pruning.
The search guarantees below concern the executable ground task.

\mypar{Returned Plans and Completion.}
In \texttt{Proofs/Search/AStar.lean}, \texttt{astar\_sound} proves that any returned plan executes from the initial state to a goal.
The theorem \texttt{astar\_optimal} compares its cost with every valid plan of cost less than $B=2^{40}$, assuming goal awareness and admissibility on an invariant containing the initial state.
Taking that invariant to be reachability gives the statement in Section~\ref{sec:method}.
The proofs do not certify dead ends, because a pruned state has no plan that costs less than~$B$ but may still have a more expensive one, even if a domain-specific dead-end test caused the pruning.
When \texttt{astar} exhausts its open list, LeanPlan therefore reports that it found no plan rather than that the task is unsolvable.
Because Lean can only reason about functions that provably terminate, \texttt{astar} takes an explicit iteration budget, its fuel, which is $2^{62}$ by default.
Both theorems hold whenever the fuel suffices to return a plan, and exhausting it is reported as a failure.
The separate theorem \texttt{astar\_terminates} in \texttt{Proofs/Search/Fuel.lean} gives a sufficient completion bound for well-formed unit-cost tasks: fuel greater than $N(N+1)+1$, where $N$ counts all packed states satisfying \texttt{State.WF t.numFacts}.

\mypar{Implementation.}
LeanPlan uses Lean~4.30.0 with mathlib at revision \texttt{c5ea00351c28e24afc9f0f84379aa41082b1188f}.
The executable and the proof library are separate build targets, \texttt{lake build planner} and \texttt{lake build Proofs}, both run from \texttt{planner/}, so checking the proofs requires the second.
The command-line interface calls the proved \texttt{astar} routine.
The library's \texttt{Proofs/AxiomAudit.lean} prints axiom dependencies for the search theorems and domain endpoints.
The permitted dependencies are Lean's standard axioms \texttt{propext}, \texttt{Classical.choice} and \texttt{Quot.sound}.
Proof-target build and audit outputs are needed to confirm this property for a particular artifact revision.

\section{New Domains}
\label{app:new-domains}

\mypar{Orrery.}
Orrery models the alignment of coupled wheels in a clockwork model of the solar system.
Each wheel has $K$ discrete angular positions, and a twist advances one wheel by one position while retreating a coupled wheel by one position, modulo $K$.
The goal aligns every wheel with its reference mark using as few twists as possible.
Every twist preserves the sum of wheel positions modulo $K$, so correcting one wheel changes another and may require routing displacement through several couplings.

\mypar{Cisterns.}
Cisterns models water redistribution through a network of hoses.
An action transfers one unit of water between connected cisterns when the source holds at least two units more than the destination.
Goals prescribe exact water levels in selected cisterns, and plans minimize the number of transfers.
Transfers conserve water but strictly decrease the sum of squared levels, making the state transitions irreversible.
Equalizing levels too early can remove the gradients needed to reach a goal.

\mypar{Qubit Routing.}
Qubit Routing models the placement of logical qubits on a hardware coupling grid while executing an ordered sequence of two-qubit gates.
A swap exchanges the logical qubits at adjacent grid nodes; a gate can execute only when its two operands occupy adjacent nodes.
The goal executes every gate in order, with no required final qubit placement.
Both swaps and gate executions have unit cost, so the fixed number of gate executions makes minimizing plan length equivalent to minimizing swaps.
The placement left after one gate affects the routing needed by later gates.

\section{Domain Heuristic Details}
\label{app:heuristics}

Table~\ref{tab:heuristic-components} summarizes the heuristics.
The entries describe how the main action-count and distance bounds combine.
The final domain theorems use the interface in Appendix~\ref{app:proofs} for all thirteen evaluators.
Distance-based terms use the checked tables described below, including their finite caps and zero-table fallback.

\begin{table}[htbp]
\centering
\caption{Components of the registered domain heuristics. A maximum combines bounds that may count the same actions. Additive terms count separate necessary actions under the domain invariants.}
\label{tab:heuristic-components}
\begin{tabular}{@{}lp{0.79\linewidth}@{}}
\toprule
Domain & Bound composition \\
\midrule
Blocksworld & Maximum of unmet-clear count and required grab/place count, with held-block corrections and two actions per forced detour. \\
Childsnack & Unserved goal children, tray-loading shortfall, sandwich-making shortfall and necessary tray moves. Shortfalls account for gluten-free demand and existing sandwiches. \\
Ferry & Unmet car goals: three actions per nonaboard car or one or two per carried car, plus the maximum of repositioning and waiting-location corrections. \\
Floortile & Unmet painted goals, color changes and the maximum of farthest painting-position distance, half the unattended-tile count rounded up and distinct forced painting squares without a robot. Color changes cover missing colors and, with one robot, the color order forced within a column. A reachability test also detects dead ends. \\
Miconic & Boarding and leaving counts plus distinct required floors other than the current lift floor. \\
Rovers & Communication, sampling, imaging, calibration-shortfall and empty-store-shortfall counts, plus the maximum of travel and unoccupied required-waypoint counts. \\
Satellite & Missing images, calibration and power-on cover bounds, possible power-slot release, missing pointing goals, unpaid image directions and a possible calibration approach turn. The cover bounds take values zero, one or two. \\
Sokoban & Maximum of total push distances plus the walking bound before the first push and the largest one-box/robot distance. \\
Spanner & Tightening actions, pickup shortfall and the distance to the gate. A usable-spanner shortage test detects dead ends. \\
Transport & Package loading and unloading actions plus the maximum of the farthest package driving requirement and distinct required locations without a vehicle. \\
\midrule
Orrery & Half the summed cyclic distances of the wheels to their goal marks, rounded up. \\
Cisterns & Maximum of two families of bounds on the level changes that goal tanks still need: demand sums over sets that no siphon can serve twice and transport bounds from tank potentials. \\
Qubit Routing & Remaining gate executions plus the maximum of the largest adjacency deficit of a remaining gate and half the summed deficits, rounded up, of remaining gates with disjoint qubits. \\
\bottomrule
\end{tabular}
\end{table}

\mypar{Blocksworld.}
For each block, compilation selects the first \texttt{on} goal naming it as the upper block and the first naming it as the lower block.
These goals specify its selected support and the block it should carry, respectively.
The goal-support chain follows the selected supports downwards.
The heuristic marks blocks with an unmet selected \texttt{on} goal or \texttt{on-table} goal, held blocks and blocks standing on a support required to be clear or to carry a different selected block.
It propagates these marks upwards through the current towers.
Each marked block contributes one grab (\texttt{pickup} or \texttt{unstack}) and one placement (\texttt{putdown} or \texttt{stack}).
The bound subtracts one grab if a marked block is already held and one placement if some marked block has no unmet support goal (\texttt{on} or \texttt{on-table}).
The latter subtraction allows such a block to remain held in the relaxation.
A marked block incurs a \emph{detour} if its selected goal support is marked and the block is either held or standing on a block that belongs below it in the compiled goal-support chain.
Such a block must be placed temporarily before it can reach its goal support, adding one grab and one placement.
The heuristic takes the maximum of this action count and the number of unmet \texttt{clear} goals.

\mypar{Childsnack.}
Let $n$ count unserved goal children and $a$ those with gluten allergies.
Let $t$ and $f$ count all sandwiches and gluten-free sandwiches on trays, and let $k$ and $g$ be the corresponding kitchen counts.
Kitchen counts exclude sandwiches already on trays.
The loading bound is $L=\max(0,a-f,n-t)$, and the sandwich-making bound is $M=\max(0,a-f-g,L-k)$.
The heuristic returns $n+L+M+R$ when $n>0$ and zero otherwise.
Here $R$ counts nonkitchen places whose waiting unserved goal children outnumber the sandwiches on trays currently there, plus one if $L>0$ and no tray is in the kitchen.
A tray without a recorded location contributes no sandwiches to any place count.
Serving, loading, making sandwiches and moving trays require separate actions, so their bounds are additive.

\mypar{Ferry.}
Each unmet car goal contributes three actions if the car is not aboard, two if it is aboard away from its destination and one if it is aboard and the ferry is at its destination.
We call a car with an unmet goal \emph{waiting} if it is not aboard, even when its location is unrecorded.
Let $W$ contain the distinct recorded locations of waiting cars.
The emptiness test reads the \texttt{empty-ferry} atom, with a missing atom counting as false.
The repositioning correction is zero if no car is waiting, if the emptiness test succeeds at a location in $W$, or if the test fails and the carried car with an unmet goal has its destination in $W$.
It is one otherwise.
The waiting-location correction counts locations in $W$ that are neither destinations of unmet car goals nor the current location of a ferry for which the emptiness test succeeds.
Both corrections bound additional actions beyond the per-car counts, which can be paid by sailing or extra car handling.
The heuristic adds the per-car counts and the maximum of these corrections.
A waiting car with no recorded location still contributes three actions, but no waiting location.
An unknown ferry location matches neither waiting locations nor destinations.

\mypar{Floortile.}
The heuristic adds the number of unmet painted goals, a color-change bound and a movement bound.
The movement bound is the maximum of three quantities.
The first takes, for each unmet painted goal, the minimum static-graph distance from any robot to any square from which the tile can be painted, then selects the largest distance.
The second counts \emph{unattended} tiles with unmet goals, which have no robot at a painting position, and divides by two, rounding up.
One move can put a robot in position to paint at most two such tiles.
The third counts distinct forced painting squares that no robot currently occupies.
A goal tile has a forced painting square below it if that square exists and the upward chain consists entirely of goal tiles and ends at a tile with nothing above it.
Since painting makes a tile impassable, these goal tiles must be painted from below, starting at the top.
The color-change bound counts the distinct required colors held by no robot.
With a single robot, it is the maximum of this count and a column bound.
Starting at an unmet tile, the column bound follows the forced squares downwards through unmet tiles and counts each change of color between consecutive tiles, plus one if the robot does not hold the first color.
The heuristic uses the largest column bound over all unmet tiles.
The heuristic also searches the movement graph through clear and robot-occupied squares.
It returns $B$ if an unmet tile's forced painting square is unreachable, or, for a tile without a forced square, if none of its painting positions is reachable.
Otherwise it returns the sum of the painting, color-change and movement bounds.

\mypar{Miconic.}
Together with the invariants, the certificate conditions in Appendix~\ref{app:task-classes} justify filtering out passengers without goals and counting each needed floor only once.
The waiting and riding sets in Section~\ref{sec:cases} partition the passengers with unmet goals, and \texttt{neededFloors} deduplicates their required floors.
An absent passenger origin contributes no floor, and an absent lift location excludes no floor from this set.
Figure~\ref{fig:miconic} shows a simplified version of the heuristic's Lean code without helper definitions.
The heuristic \texttt{improved} calls \texttt{compile} once per task to build passenger and floor tables and evaluates \texttt{value} in each state.
\begin{figure}[t]
\begin{lstlisting}[language=Lean]
def value (d : Data) (s : State) : Nat :=
  2 * (waiting d s).size + (riding d s).size
    + (neededFloors d s).length

def improved (t : Task) : Heuristic :=
  let d := compile t
  { name := "miconic-improved", eval := value d }
\end{lstlisting}
\caption{Miconic's heuristic counts waiting and riding passengers and required floors other than the current floor. Helper definitions are omitted.}
\label{fig:miconic}
\end{figure}

\mypar{Rovers.}
The heuristic counts one communication per unmet data goal, one sample per missing soil or rock analysis and one image-taking action per missing image.
An analysis or image is missing only if no rover already holds it.
Each image-taking action consumes a calibration, so the calibration shortfall is the number of required image-taking actions minus the number of currently true \texttt{calibrated} atoms, truncated at zero.
Similarly, the empty-store shortfall is the number of required samples minus the number of currently true \texttt{empty} atoms, truncated at zero.
The travel bound is the largest distance to a required sampling waypoint from its nearest suitably equipped rover, using each rover's traversal-and-visibility graph.
Rovers without recorded locations are omitted, and a minimum with no located suitable rover is zero.
A second navigation bound counts distinct required sampling waypoints that no rover currently occupies.
The heuristic adds the five action counts to the maximum of the two navigation bounds.
Communication, sampling, imaging, calibration, emptying stores and navigation use separate actions.

\mypar{Satellite.}
Each unmet \texttt{have\_image} or \texttt{pointing} goal contributes one.
The heuristic adds the following five bounds to these counts.
An instrument is \emph{ready} if it is both powered and calibrated.
The calibration bound considers modes needed by unmet image goals that no ready instrument supports.
It is zero if no mode remains, one if a single instrument supports all remaining modes and two otherwise.
The power-on bound first removes modes supported by a powered instrument, then applies the same rule to the remaining modes, considering all instruments for the cover.
For each unmet image, we omit its direction if an unmet pointing goal already names it.
We also omit it if a satellite points there with an instrument that supports the image's mode and is either ready or has that direction as its calibration target.
The image-direction bound counts distinct directions of the remaining images.
For the remaining terms, a candidate is any instrument that supports a mode needed by an unmet image goal.
The calibration-approach bound is zero if no calibration is needed.
It is also zero if any candidate has an unknown carrier or target, its carrier is already at its target, or its target occurs among the unmet pointing-goal directions or additional image directions.
Otherwise it is one.
The power-slot-release bound is one if some needed mode has no powered instrument and every candidate has a known carrier with a recorded, currently false \texttt{power\_avail} atom.
Otherwise it is zero.

\mypar{Sokoban.}
The one-box product graph behind $J_b(s)$ keeps the robot because reaching a pushing side may require walking even when a box is close to its goal.
The runtime registration selects \texttt{baseValue}, which computes the maximum from Section~\ref{sec:cases}.
A separate function \texttt{value} adds an explicit dead-end branch, but the planner does not use this wrapper.
The implementation uses finite caps for unreachable distance-table entries and checks the properties required by the proofs.
For $n$ locations, an unrecorded box position contributes $n$ to its push bound and $n^2$ to its product bound when a robot position is recorded.
An unrecorded robot position makes the product and walking bounds zero.
The walking bound is the table cap if no recorded box location has a relaxed pushing position.
All components are zero when no compiled box-location goal remains unmet, even if another dynamic goal is still unmet.

\mypar{Spanner.}
The implementation permits several workers.
Let $G$ contain the goal nuts with a recorded initial location.
The compiled heuristic omits other goal nuts because nut positions never change and an unlocated nut cannot be tightened.
The movement bound $D(s)$ is the directed-link distance from the closest worker to the gate.
The implementation computes it as the largest such distance to the location of a nut in $G$ with an unmet \texttt{tightened} goal, which is the gate in every IPC 2023 task, and it is zero if no such nut exists.
Workers without a recorded location are omitted from the minimum over workers, which equals the distance-table cap if no recorded worker location remains.
Unreachable distances use the finite bound stored in the compiled distance table.
The ground-spanner count excludes carried spanners and counts only usable objects at a location reachable from some worker.

\mypar{Transport.}
Each unmet package goal contributes two handling actions if the package is on the ground and one if it is carried.
For a ground package, the driving requirement adds the shortest road distance from any vehicle to the package and the distance from there to its destination.
For a carried package, it uses the carrying vehicle's distance to the destination.
The first driving bound is the maximum requirement over packages, with an empty maximum equal to zero.
The second counts distinct destinations and recorded ground locations of packages with unmet goals that currently contain no vehicle.
The heuristic adds the handling count and the maximum of these driving bounds.
A package with neither a ground location nor a carrier contributes zero handling and individual driving cost, but its destination remains in the location count.
Missing vehicle locations use the distance-table cap.

\mypar{Orrery.}
Each goal fixes the mark of one wheel.
Compilation builds the cycle of marks from the static \texttt{succ} atoms and adds the reverse edges, since a twist advances one wheel and retreats another.
For each goal wheel~$w$, let $\delta_w(s)$ be the distance in this symmetric cycle from its current mark to its goal mark.
A twist moves two wheels by one mark each, so it reduces $\sum_w\delta_w(s)$ by at most two, and the heuristic is $\lceil\sum_w\delta_w(s)/2\rceil$.
If a state assigns several marks to a wheel, $\delta_w(s)$ uses the closest one, and a wheel without a mark contributes the distance-table cap.
The heuristic does not exploit that twists conserve the sum of wheel positions modulo~$K$ (Appendix~\ref{app:new-domains}).

\mypar{Cisterns.}
Each goal fixes the level of one tank, which we call a goal tank.
For a goal tank~$t$, let $e_t(s)$ be its current level minus its goal level, so that $t$ must still lose $\max(0,e_t(s))$ units and gain $\max(0,-e_t(s))$ units.
We call these quantities the \emph{loss} and \emph{gain demands} of~$t$.
A siphon from $u$ to $v$ lowers $u$ and raises $v$ by one level, so it reduces at most the loss demand of $u$ and the gain demand of~$v$, each by one.
The heuristic is the maximum of two families of lower bounds, both fixed once per task during compilation.

The first family considers sets of demands in which no loss demand belongs to a tank that shares a hose with the tank of a gain demand.
A siphon reduces the sum of such a set by at most one, so the sum is a lower bound.
Compilation selects 28 such sets greedily.
It scans the demands in fixed orders, in orders of decreasing initial demand and in orders of decreasing ratio of initial demand to the number of conflicting demands.
Eight sets instead contain both demands of every tank in a set of goal tanks of which no two share a hose.

The second family uses integer \emph{potentials}~$\pi$ on the goal tanks, extended by zero to all other tanks, such that the potentials of two tanks sharing a hose differ by at most one.
A siphon from $u$ to $v$ changes $\sum_t\pi_t e_t(s)$ by $\pi_v-\pi_u$, so $\left|\sum_t\pi_t e_t(s)\right|$ is a lower bound.
Such potentials are dual solutions of a transportation problem that ships excess water along the hoses to the goal tanks that lack it, with all other tanks acting as one reservoir.
Compilation computes 17 candidate potentials, 16 by coordinate-wise improvement in different tank orders and one from a maximum-weight closure problem solved by maximum flow.
It keeps every valid candidate with the largest bound in the initial state.
Both families are zero when all goal tanks are at their goal levels.
As with the distance tables, the proofs do not rely on the greedy selection or the flow computation, because the certificate checks the resulting sets and potentials against every ground action (Appendix~\ref{app:task-classes}).

\mypar{Qubit Routing.}
The static \texttt{succ} atoms order the gates in a chain that ends at the goal gate.
For each gate, compilation records the number of gates from it up to the goal gate, each of which requires one \texttt{execute} action.
The heuristic adds this count for the active gate to a bound on the remaining swaps.
The \emph{deficit} of a gate is the distance between the nodes of its two qubits in the coupling graph minus one, which is zero if the qubits are adjacent.
A swap moves two qubits by one edge each, so it reduces the deficit of any gate by at most one, and the largest deficit among the remaining gates is a lower bound on the number of swaps.
If several remaining gates have pairwise disjoint qubits, a swap moves at most two of their qubits and reduces their summed deficits by at most two, so half this sum, rounded up, is also a lower bound.
Compilation fixes $2m$ such sets for the $m$ gates of the chain by greedy maximal matching, scanning the gates in each of the $m$ rotations of the chain order, forward and backward.
In each state, the heuristic restricts these sets to the remaining gates.
The swap bound is the maximum of the largest deficit and the matching bounds.
Since executing a gate requires adjacent qubits, it removes a gate with zero deficit and reduces the heuristic by exactly one.

\subsection{Checked Task Classes}
\label{app:task-classes}

The following certificate conditions delimit the supported task classes.
They apply in addition to ordinary validation and exact equality of the action-schema list with the generated domain.
An \emph{exact type} requires each object that inhabits that type through subtyping to declare that precise type.
The initial-state checks establish the premises whose dynamic components the domain proofs preserve.
Static-predicate checks ensure that the atoms in their static components cannot change.
A dynamic goal uses a predicate that an action schema can change.
These are task acceptance conditions, distinct from the zero-table fallback used by distance computations.
The definitions are in each domain's \texttt{Certificate.lean} file.

\mypar{Blocksworld.}
The certificate requires a declared \texttt{object} type, at least one object and duplicate-free goals.
Each \texttt{on} goal has two arguments, but goals need not consist only of \texttt{on} atoms.
Initially, each block has at most one support and supports at most one block.
At most one block is held, and \texttt{arm-empty} holds exactly when none is held.
A held block neither stands on nor supports a block and is neither on the table nor clear.
A block standing on another is not on the table, and a clear block supports nothing.
Following initial supports must terminate within the number of \texttt{on} atoms plus one.
The checks establish unambiguous acyclic towers without requiring complete block locations.

\mypar{Childsnack.}
The sandwich, tray, place, child, bread-portion and content-portion types must be declared, and sandwich, tray and place types must be exact.
At least one tray and sandwich and a place named \texttt{kitchen} must exist.
Goals form a nonempty, duplicate-free list of \texttt{served} atoms naming children declared with type \texttt{child}.
Each goal child has a waiting place and an allergy classification.
Waiting places are unique per child and typed, and allergic and nonallergic classifications are disjoint.
Both classification predicates and \texttt{waiting} are static.
Initially, each tray occupies at most one place, each sandwich belongs to at most one tray, and kitchen and tray membership are disjoint.
Nonexistent sandwiches are in neither location and cannot be gluten-free.
No sandwich is initially gluten-free.
Tray-location existence is not required.
These checks support distinct resource counts and preserved sandwich and tray invariants.

\mypar{Ferry.}
Car and location types must be declared and exact, with at least one object of each.
Every car has exactly one \texttt{at} goal naming a typed location.
The extracted car--destination pairs and their car names are duplicate-free, but other goal predicates are not excluded.
Initially, the ferry has at most one location, each car has at most one ashore location, and at most one car is aboard.
An aboard car has no ashore location.
An empty-ferry atom implies that no car is aboard, without requiring the converse.
Initial location existence is not required.
These conditions establish the capacity-one invariant used by the per-car sailing bound.

\mypar{Floortile.}
Tile, robot and color types are exact.
Goals are duplicate-free, name each painted tile once and exclude grid edges.
Grid predicates are static.
Initially, each robot has at most one position and color, and each tile has at most one robot.
Occupied tiles are neither clear nor painted, and clear tiles are unpainted.
The certificate grounds both relevance settings and checks each compiled task.
Up/down edges have arity two, \texttt{up} is functional and inverse-functional, and each \texttt{down} edge reverses an \texttt{up} edge.
Compiled painted goals have known colors and unique tile entries, and present clear-atom indices match their tiles.
For $N$ compiled tile entries and distance cap $D_{\mathrm{bound}}$, the certificate requires $2N+\max(D_{\mathrm{bound}},N)\leq B$.
Neither robot-position existence nor complete rectangular topology is required.

\mypar{Miconic.}
Floor and passenger types must be declared, with at least one object of each, and the floor type must be exact.
Goals are a nonempty list of \texttt{served} atoms naming distinct passengers declared with type \texttt{passenger}.
Destinations are static, unique per passenger, typed as floors and present for every goal passenger.
The initial invariant checks at most one lift floor and at most one origin per passenger.
These multiplicity checks do not require that either location exists.

\mypar{Rovers.}
Rover, waypoint, store, camera, mode, lander and objective types must be declared, and rover and waypoint types must be exact.
Goals are duplicate-free and dynamic, but need not be communication goals.
The heuristic compiles only the three communicated-data goal predicates.
Each rover initially has at most one position, without requiring one to exist.
Traversal, visibility and soil/rock equipment predicates must be static.
Traversal self-loops are forbidden.
Every traversal edge also present in \texttt{visible} must have reverse traversal and visibility atoms.
The per-rover graphs intersect traversal and visibility.
There are no additional explicit type-disjointness, store/camera assignment-multiplicity or task-size conditions.
Graph distances use node-count caps and checked zero-table fallback.

\mypar{Satellite.}
Satellite, direction, instrument and mode types must be declared and exact.
Goals are duplicate-free and dynamic, but need not be image or pointing goals, the only forms compiled into heuristic goal tables.
Initial instrument-carrier, calibration-target and satellite-pointing relations are single-valued on their first argument, without requiring any entry to exist.
Initial \texttt{pointing} and \texttt{not-pointing} atoms must not contradict each other.
Every instrument assigned to a carrier initially lacks \texttt{power\_on}.
The \texttt{on\_board}, \texttt{supports} and \texttt{calibration\_target} predicates are static.
The certificate does not require complete pointing complements, initial power availability or complete relation coverage, and imposes no task-size condition.

\mypar{Sokoban.}
Location, direction and box types must be declared, and location and box types must be exact.
Goals are duplicate-free and dynamic, with at most one \texttt{at} destination per box.
Other dynamic goals are permitted, and neither a goal nor an initial location is required for every box.
The robot initially occupies exactly one square, which is clear.
Each box initially occupies at most one square, each square contains at most one box, and box-occupied squares are not clear.
Static adjacency forbids self-loops and requires reverse edges with exactly typed direction objects.
For $n$ locations, checked push and walk distances use cap $n$, and product distances use cap $n^2$, with zero-table fallback.
There is no certificate bound relative to $B$.

\mypar{Spanner.}
Location, man and spanner types must be exact.
Nut, man and spanner objects must inhabit pairwise disjoint types.
Links are static, and goals are duplicate-free and contain no links.
Other goal predicates are not excluded.
Initially, each object has at most one location, no nut is both loose and tightened and a carried object has no ground location.
There is no requirement that every object has a location or that only one worker exists.
The compiled-task bound is $2|G|+D_{\mathrm{bound}}\leq B$ for the set $G$ defined above, where $D_{\mathrm{bound}}$ is the distance-table bound.

\mypar{Transport.}
Location and vehicle types are exact, loading and unloading parameter types are declared and no vehicle is also a package.
Goals are duplicate-free, exclude \texttt{road} and contain at most one \texttt{at} destination per object, with no such goal naming a vehicle.
Other goal predicates are permitted.
Initially, each object has at most one location and at most one carrier, and cannot have both.
Neither location nor carrier existence is required.
Road and capacity-predecessor predicates are static, without road symmetry, connectivity or capacity-multiplicity checks.
Road distances use checked node-count caps and zero-table fallback.
The certificate imposes no size bound relative to $B$.

\mypar{Orrery.}
Every compiled goal names a task goal and a mark of the cycle.
Every ground action adds marks of at most two goal wheels, and each added mark is one cycle step away from a mark of the same wheel in the action's precondition.
The certificate imposes no initial-state conditions, since the heuristic takes the closest of a wheel's marks.

\mypar{Cisterns.}
Every compiled goal names a task goal and a level of the level graph, and the level atoms of each goal tank are distinct.
Each goal tank has exactly one initial level, and every ground action preserves the mutual exclusion of its level atoms.
Every ground action either leaves a goal tank unchanged or replaces the level in its precondition by one new level.
Demands number the levels in the order of the task's level objects, and each action reduces every demand by at most one.
Each action reduces every compiled set sum by at most its cost and changes every potential bound by at most its cost.

\mypar{Qubit Routing.}
The goal is a single \texttt{active} atom whose gate ends the compiled chain, and the two qubits of each gate differ.
Exactly one gate is initially active, and every ground action preserves the mutual exclusion of the \texttt{active} atoms.
Likewise, each qubit initially occupies exactly one node, and every ground action preserves the mutual exclusion of its position atoms.
An action that changes the active gate activates its successor, moves no qubit and requires the qubits of the executed gate to be at distance at most one.
Every action moves at most two qubits, and the gates of each compiled matching have pairwise disjoint qubits.
For every gate, each action reduces the deficit by at most one and by at most the number of the gate's qubits that it moves.

\mypar{Distance-Table Checks.}
The shared graph-distance check verifies table dimension bounds, bounded entries, a zero diagonal, edge consistency and a cap at least as large as the node count.
It also checks the following property for every edge from $u$ to $v$ and every target $t$: if the entry from $v$ to $t$ is below the cap, then the entry from $u$ to $t$ is also below the cap.
Sokoban's product check requires $n^2$ entries, zero at valid goal configurations, bounded entries, edge consistency and the same backward-propagation property.
These checks connect the finite tables to the distance properties used by the domain proofs.
Failure selects zero tables during compilation rather than rejecting the task.

\section{Experiment Records}
\label{app:records}

The dataset contains all 8\,550 recorded configuration--task runs, including the 5\,850 runs of the five configurations reported here, with coverage, cost, search and total time, expansions, resource limits, certificate flags and whether expansions came from an interrupted run's checkpoint.
The IPC 2023 snapshot contains 7\,200 runs, and a separate snapshot adds 1\,350 runs on the new domains, retaining their original algorithm identifiers.
Every reported configuration has a run for each of the 90 test tasks per domain, and all 1\,157 solved runs of the reported configurations carry expansion measurements from completed runs.
We exclude checkpoint counts from the expansion comparisons.

On IPC 2023, the three-way intersection of the heuristic configurations has 198 tasks, the LeanPlan--SCP pair 211 and the pair of blind configurations 154.
Figure~\ref{fig:performance} uses this suite only.
The runtime curves use \texttt{total\_time} for every configuration.
Only solved runs contribute, so each curve ends at full-budget coverage.
We draw times below one millisecond at one millisecond on the logarithmic axis, whose values are per-task thresholds rather than a cumulative budget.

For LeanPlan, total time is wall-clock time from before PDDL reading through search, covering validation, grounding, $h^2$ pruning and heuristic setup, and excluding plan output and external VAL checking.
Search time runs from the end of grounding and $h^2$ pruning until search returns, including certificate checking and heuristic setup.
LeanPlan counts a state as expanded when it generates its successors, excluding stale queue entries and the popped goal.
Each planner reports expansions under its own definition.

\section{Agent Prompt}
\label{app:prompt}

The template below reproduces the prompt text from the agentic-loop implementation, with angle-bracket placeholders for the fields filled by the controller.
These include domain and file names, training tasks, baseline results, proof endpoints, certificate instructions and worked examples.
We render the function arrow as \texttt{->} and standardize the spelling of $h^2$-preprocessing; the instructions are otherwise unchanged.

\begin{lstlisting}[basicstyle=\fontsize{8}{10}\selectfont\ttfamily,breaklines=true,
  breakatwhitespace=true,breakautoindent=false,breakindent=1em,
  columns=fullflexible,frame=none,aboveskip=8pt,belowskip=8pt]
<role>
You are a highly-skilled AI planning researcher and a proficient
Lean 4 programmer.  You work in the LeanPlan repository at `<repo>`,
an optimal STRIPS planner written in Lean 4.  Write a
domain-dependent heuristic for the PDDL domain
<domain><domain-name></domain>, and a Lean proof that it is
admissible.  A* stays optimal only if the heuristic is admissible,
and Lean's kernel decides whether it is.
</role>

<objective>
Create the registry heuristic `<heuristic-name>` with the best
possible performance on the <training-count> training tasks in
`<training-dir>`. Make it as informed and domain-dependent as
possible, and prove it admissible. Test and refine it on the
training tasks. Beating the Scorpion baseline is the minimum
requirement. Once you beat it, you must continue refining the
heuristic until you cannot improve it further.

The baseline is Scorpion with h^2-preprocessing and the search
configuration below. It uses the documented Scorpion configuration
with three time limits changed: 1 second to build the Sys-SCP
pattern collection, 1 second per restart, and 15 seconds of
cumulative online order-finding during search. Your heuristic must
be goal-aware and consistent on every reachable state its
certificate admits. Goal-awareness and consistency give
admissibility, which gives optimality; the shared library derives
that step for you.

<baseline-results>

Both planners get the same budget per task, enforced the same way
with `runlim` on one pinned core: <time-limit> seconds of wall-clock
time, one CPU core.

You work alone until you finish.  Nobody will answer you, and nobody
will prompt you again. Run the checks below yourself and act on what
they say.
</objective>

<what-you-produce>
<output-file-count>

1. The runtime heuristic, `<runtime-path>`, in namespace
   `<runtime-namespace>`, exporting `<runtime-function> : Task ->
   Heuristic` whose `name` field is exactly `<heuristic-name>`.  It
   compiles against `Planner/` alone, so it cannot import mathlib or
   anything under `Proofs/`.

2. The proof, `<proof-path>`, module `<proof-module>`, declaring
   these endpoints:
<theorem-endpoints>

<certificate-instructions>

Add the registry entry for `<heuristic-name>` in
`planner/Planner/ExampleHeuristics/Registry.lean`, with the imports
it needs, so that `--heuristic <heuristic-name>` runs.  Until you
do, every check below reports an unknown heuristic.
</what-you-produce>

<success-criteria>
Success means finding the best admissible heuristic you can for the
training set. Beating the baseline is only the minimum. We expect
you to push performance as far beyond that minimum as you can
through continued refinement.

The final candidate must satisfy these correctness requirements:

1. Goal-aware: the heuristic is 0 on every goal state the
   certificate admits.
2. Consistent: `h s <= cost o + h (o s)` for every grounded operator
   `o` and every reachable state `s` the certificate admits.
3. The theorems concern the compiled heuristic the registry builds,
   which is the function A* evaluates, not an abstract surrogate.
4. The certificate accepts all <training-count> training tasks.
</success-criteria>

<how-to-check-your-own-work>
Run these from `<repo>`.

Every planner command below is wrapped in `runlim`, at the same
<time-limit>-second limit the controller uses.  Run them that way.
A command that would not finish inside the limit is a failure, and
you want to see it as one rather than wait for it.

```bash
# 1. The runtime heuristic compiles.  Seconds, and no mathlib.
(cd planner && lake build planner)

# 2. Walk reachable states and report any state where the heuristic is not
#    goal-aware or not consistent, naming the transition that breaks it.  A
#    breach here means no proof exists.
runlim --real-time-limit=<time-limit> --kill --propagate \
    <planner-binary> --heuristic <heuristic-name> \
    --audit-consistency 20000 <domain-file> <first-training-task>

# 3. The certificate accepts a task.  The controller runs this on every task
#    under the same limit, so a certificate too slow to finish fails.
runlim --real-time-limit=<time-limit> --kill --propagate \
    <planner-binary> --heuristic <heuristic-name> \
    --check-certificate <domain-file> <first-training-task>

# 4. The proof.  Minutes; this is the slow one.
(cd planner && lake build <proof-module>)

# 5. Write a scratch file importing <proof-module> that runs
#    `#print axioms` on each endpoint, then elaborate it:
(cd planner && lake env lean /tmp/audit.lean)

# 6. What the search actually does on a task.
runlim --real-time-limit=<time-limit> --kill --propagate \
    <planner-binary> --heuristic <heuristic-name> <domain-file> <middle-training-task>

# 7. The same task under Scorpion, for comparison.  Scorpion lives at
#    $SCORPION_SRC, and gets the same budget: <time-limit> s, one core.
runlim --real-time-limit=<time-limit> --kill --propagate \
    taskset --cpu-list 0 "$SCORPION_SRC/fast-downward.py" \
    --preprocess <domain-file> <middle-training-task> \
    --search "<scorpion-search>"
```
</how-to-check-your-own-work>

<what-not-to-do>
- Do not write `sorry`, `admit`, or `native_decide`, and do not
  declare an `axiom`.  The controller scans for all four before it
  checks anything else.
- Do not weaken a required theorem to make it easy.  Each endpoint
  states admissibility of the compiled heuristic under the
  certificate.  A theorem that is true but says less fails.
- Do not use the certificate to dodge difficulty.  It states what
  your proof assumes; the controller runs it on every training task
  and rejects a candidate that refuses one.
</what-not-to-do>

<when-to-stop>
Stop when you are confident that the heuristic is as informed as you
can make it and you cannot refine it further. Leave the best
candidate, with a working proof and passing certificates, in the
files.
</when-to-stop>

<orientation>
Read these first, in roughly this order:

<reading-list>

<worked-example-description>

The domain file is `<domain-file>`, and the <training-count>
training tasks are in `<training-dir>`, for example
<example-training-tasks>.  The registry name `<heuristic-name>` maps
to the PDDL domain `<pddl-domain-name>`.  The planner binary is
`<planner-binary>`.
</orientation>
\end{lstlisting}

%% file: figure-pddl.tex
\begin{figure}[htbp]
\centering
\begin{subfigure}[t]{0.53\linewidth}
\captionsetup{position=top}
\caption{Miconic domain excerpt.}
\begin{lstlisting}[language=PDDL,basicstyle=\footnotesize\ttfamily,breaklines=true]
(define (domain miconic)
  (:requirements :strips :typing)
  (:types passenger - object
          floor - object)
  ; Predicate declarations omitted.
  (:action board
    :parameters (?f - floor
                 ?p - passenger)
    :precondition (and (lift-at ?f)
                       (origin ?p ?f))
    :effect (and (boarded ?p)
                 (not (origin ?p ?f))))
  ; Other actions omitted.
)
\end{lstlisting}
\end{subfigure}\hfill
\begin{subfigure}[t]{0.44\linewidth}
\captionsetup{position=top}
\caption{Miconic task \texttt{easy-p01.pddl}.}
\begin{lstlisting}[language=PDDL,basicstyle=\footnotesize\ttfamily,breaklines=true]
(define (problem miconic-01)
  (:domain miconic)
  (:objects p1 - passenger
            f1 f2 f3 f4 - floor)
  (:init (lift-at f1)
    (origin p1 f2)
    (destin p1 f3)
    (above f1 f2) (above f1 f3)
    (above f1 f4) (above f2 f3)
    (above f2 f4) (above f3 f4))
  (:goal (and (served p1))))
\end{lstlisting}
\end{subfigure}
\caption{PDDL from the Miconic benchmark used in our experiments.
The domain excerpt shows the boarding action and omits predicate declarations and other actions.
The complete task asks the lift to carry passenger \texttt{p1} from floor \texttt{f2} to \texttt{f3}.}
\label{fig:pddl}
\end{figure}

%% file: paper.bbl
\begin{thebibliography}{73}
\providecommand{\natexlab}[1]{#1}
\providecommand{\url}[1]{\texttt{#1}}
\expandafter\ifx\csname urlstyle\endcsname\relax
  \providecommand{\doi}[1]{doi: #1}\else
  \providecommand{\doi}{doi: \begingroup \urlstyle{rm}\Url}\fi

\bibitem[Abate et~al.(2018)Abate, David, Kesseli, Kroening, and
  Polgreen]{abate-cav2018}
Alessandro Abate, Cristina David, Pascal Kesseli, Daniel Kroening, and
  Elizabeth Polgreen.
\newblock Counterexample guided inductive synthesis modulo theories.
\newblock In \emph{Proc.\ CAV 2018, Part I}, pp.\  270--288, 2018.

\bibitem[Abdulaziz \& Kurz(2023)Abdulaziz and Kurz]{abdulaziz-kurz-aaai2023}
Mohammad Abdulaziz and Friedrich Kurz.
\newblock Formally verified {SAT}-based {AI} planning.
\newblock In \emph{Proc.\ {AAAI} 2023}, pp.\  14665--14673, 2023.

\bibitem[Abdulaziz \& Lammich(2018)Abdulaziz and
  Lammich]{abdulaziz-lammich-ictai2018}
Mohammad Abdulaziz and Peter Lammich.
\newblock A formally verified validator for classical planning problems and
  solutions.
\newblock In \emph{Proc.\ ICTAI 2018}, pp.\  474--479, 2018.

\bibitem[Abdulaziz et~al.(2022)Abdulaziz, Pommerening, and
  Corr{\^e}a]{abdulaziz-et-al-icaps2022wshsdip}
Mohammad Abdulaziz, Florian Pommerening, and Augusto~B. Corr{\^e}a.
\newblock Mechanically proving guarantees of generalized heuristics: First
  results and ongoing work.
\newblock In \emph{ICAPS Workshop on Heuristics and Search for
  Domain-independent Planning}, 2022.

\bibitem[Aggarwal et~al.(2025)Aggarwal, Parno, and
  Welleck]{aggarwal-et-al-icml2025}
Pranjal Aggarwal, Bryan Parno, and Sean Welleck.
\newblock {AlphaVerus}: Bootstrapping formally verified code generation through
  self-improving translation and {Treefinement}.
\newblock In \emph{Proc.\ ICML 2025}, pp.\  587--615, 2025.

\bibitem[Agostinelli et~al.(2019)Agostinelli, McAleer, Shmakov, and
  Baldi]{agostinelli-et-al-nmi2019}
Forest Agostinelli, Stephen McAleer, Alexander Shmakov, and Pierre Baldi.
\newblock Solving the {Rubik}'s cube with deep reinforcement learning and
  search.
\newblock \emph{Nature Machine Intelligence}, 1:\penalty0 356--363, 2019.
\newblock \doi{10.1038/s42256-019-0070-z}.

\bibitem[{Alc\'azar} \& Torralba(2015){Alc\'azar} and
  Torralba]{alcazar-torralba-icaps2015}
Vidal {Alc\'azar} and {\'A}lvaro Torralba.
\newblock A reminder about the importance of computing and exploiting
  invariants in planning.
\newblock In \emph{Proc.\ ICAPS 2015}, pp.\  2--6, 2015.

\bibitem[Arfaee et~al.(2011)Arfaee, Zilles, and Holte]{arfaee-et-al-aij2011}
Shahab~J. Arfaee, Sandra Zilles, and Robert~C. Holte.
\newblock Learning heuristic functions for large state spaces.
\newblock \emph{Artificial Intelligence}, 175:\penalty0 2075--2098, 2011.

\bibitem[Asai \& Fukunaga(2016)Asai and Fukunaga]{asai-fukunaga-aaai2016}
Masataro Asai and Alex Fukunaga.
\newblock Tiebreaking strategies for {$A^*$} search: How to explore the final
  frontier.
\newblock In \emph{Proc.\ {AAAI} 2016}, pp.\  673--679, 2016.

\bibitem[Asai \& Fukunaga(2017)Asai and Fukunaga]{asai-fukunaga-jair2017}
Masataro Asai and Alex Fukunaga.
\newblock Tie-breaking strategies for cost-optimal best first search.
\newblock \emph{Journal of Artificial Intelligence Research}, 58:\penalty0
  67--121, 2017.

\bibitem[Behnke et~al.(2026)Behnke, Kilian, and
  Gattinger]{behnke-et-al-icaps2026wshsdip}
Gregor Behnke, Simone Kilian, and Malvin Gattinger.
\newblock {$A^*$} with {$h^{\max}$} definitely finds optimal plans -- formally
  verifying a planner based on heuristic search.
\newblock In \emph{ICAPS Workshop on Heuristics and Search for
  Domain-independent Planning}, 2026.

\bibitem[Bonet \& Geffner(2001)Bonet and Geffner]{bonet-geffner-aij2001}
Blai Bonet and H{\'e}ctor Geffner.
\newblock Planning as heuristic search.
\newblock \emph{Artificial Intelligence}, 129\penalty0 (1):\penalty0 5--33,
  2001.

\bibitem[Chen et~al.(2024{\natexlab{a}})Chen, Thi{\'e}baux, and
  Trevizan]{chen-et-al-aaai2024}
Dillon~Z. Chen, Sylvie Thi{\'e}baux, and Felipe Trevizan.
\newblock Learning domain-independent heuristics for grounded and lifted
  planning.
\newblock In \emph{Proc.\ {AAAI} 2024}, pp.\  20078--20086, 2024{\natexlab{a}}.

\bibitem[Chen et~al.(2024{\natexlab{b}})Chen, Trevizan, and
  Thi{\'e}baux]{chen-et-al-icaps2024}
Dillon~Z. Chen, Felipe Trevizan, and Sylvie Thi{\'e}baux.
\newblock Return to tradition: Learning reliable heuristics with classical
  machine learning.
\newblock In \emph{Proc.\ ICAPS 2024}, pp.\  68--76, 2024{\natexlab{b}}.

\bibitem[Corr{\^e}a et~al.(2025{\natexlab{a}})Corr{\^e}a, Pereira, and
  Seipp]{correa-et-al-arxiv2025}
Augusto~B. Corr{\^e}a, Andr{\'e}~G. Pereira, and Jendrik Seipp.
\newblock Frontier large language models rival state-of-the-art planners.
\newblock arXiv:2511.09378 [cs.AI], 2025{\natexlab{a}}.

\bibitem[Corr{\^e}a et~al.(2025{\natexlab{b}})Corr{\^e}a, Pereira, and
  Seipp]{correa-et-al-neurips2025}
Augusto~B. Corr{\^e}a, Andr{\'e}~G. Pereira, and Jendrik Seipp.
\newblock Classical planning with {LLM}-generated heuristics: Challenging the
  state of the art with {Python} code.
\newblock In \emph{Proc.\ NeurIPS 2025}, pp.\  42070--42108,
  2025{\natexlab{b}}.

\bibitem[de~Moura \& Ullrich(2021)de~Moura and Ullrich]{moura-ullrich-cade2021}
Leonardo de~Moura and Sebastian Ullrich.
\newblock The {Lean} 4 theorem prover and programming language.
\newblock In \emph{Proc.\ CADE 2021}, pp.\  625--635, 2021.

\bibitem[Dold et~al.(2025)Dold, Helmert, Nordstr{\"o}m, R{\"o}ger, and
  Schindler]{dold-et-al-icaps2025}
Simon Dold, Malte Helmert, Jakob Nordstr{\"o}m, Gabriele R{\"o}ger, and Tanja
  Schindler.
\newblock Pseudo-{Boolean} proof logging for optimal classical planning.
\newblock In \emph{Proc.\ ICAPS 2025}, pp.\  54--63, 2025.

\bibitem[Edelkamp(2001)]{edelkamp-ecp2001}
Stefan Edelkamp.
\newblock Planning with pattern databases.
\newblock In \emph{Proc.\ ECP 2001}, pp.\  84--90, 2001.

\bibitem[Eriksson et~al.(2017)Eriksson, R{\"o}ger, and
  Helmert]{eriksson-et-al-icaps2017}
Salom{\'e} Eriksson, Gabriele R{\"o}ger, and Malte Helmert.
\newblock Unsolvability certificates for classical planning.
\newblock In \emph{Proc.\ ICAPS 2017}, pp.\  88--97, 2017.

\bibitem[Eriksson et~al.(2018)Eriksson, R{\"o}ger, and
  Helmert]{eriksson-et-al-icaps2018}
Salom{\'e} Eriksson, Gabriele R{\"o}ger, and Malte Helmert.
\newblock A proof system for unsolvable planning tasks.
\newblock In \emph{Proc.\ ICAPS 2018}, pp.\  65--73, 2018.

\bibitem[Ernandes \& Gori(2004)Ernandes and Gori]{ernandes-gori-ecai2004}
Marco Ernandes and Marco Gori.
\newblock Likely-admissible and sub-symbolic heuristics.
\newblock In \emph{Proc.\ ECAI 2004}, pp.\  613--617, 2004.

\bibitem[Ferber et~al.(2020)Ferber, Helmert, and
  Hoffmann]{ferber-et-al-ecai2020}
Patrick Ferber, Malte Helmert, and J{\"o}rg Hoffmann.
\newblock Neural network heuristics for classical planning: A study of
  hyperparameter space.
\newblock In \emph{Proc.\ ECAI 2020}, pp.\  2346--2353, 2020.

\bibitem[Fikes \& Nilsson(1971)Fikes and Nilsson]{fikes-nilsson-aij1971}
Richard~E. Fikes and Nils~J. Nilsson.
\newblock {STRIPS}: {A} new approach to the application of theorem proving to
  problem solving.
\newblock \emph{Artificial Intelligence}, 2:\penalty0 189--208, 1971.

\bibitem[First et~al.(2023)First, Rabe, Ringer, and Brun]{first-et-al-fse2023}
Emily First, Markus~N. Rabe, Talia Ringer, and Yuriy Brun.
\newblock {Baldur}: Whole-proof generation and repair with large language
  models.
\newblock In \emph{Proc.\ ESEC/FSE 2023}, pp.\  1229--1241, 2023.

\bibitem[Franc{\`e}s et~al.(2019)Franc{\`e}s, Corr\^{e}a, Geissmann, and
  Pommerening]{frances-et-al-ijcai2019}
Guillem Franc{\`e}s, Augusto~B. Corr\^{e}a, Cedric Geissmann, and Florian
  Pommerening.
\newblock Generalized potential heuristics for classical planning.
\newblock In \emph{Proc.\ IJCAI 2019}, pp.\  5554--5561, 2019.

\bibitem[Futuhi \& Sturtevant(2026)Futuhi and
  Sturtevant]{futuhi-sturtevant-iclr2026}
Ehsan Futuhi and Nathan~R. Sturtevant.
\newblock Learning admissible heuristics for {A}$^*$: Theory and practice.
\newblock In \emph{Proc.\ ICLR 2026}, 2026.

\bibitem[Gestrin \& Seipp(2026)Gestrin and
  Seipp]{gestrin-seipp-icaps2026wslm4plan}
Elliot Gestrin and Jendrik Seipp.
\newblock {LLM}-evolved domain-independent heuristics for symbolic {AI}
  planning.
\newblock In \emph{ICAPS Workshop on Planning in the Era of {LLMs}}, 2026.

\bibitem[Ghallab et~al.(2004)Ghallab, Nau, and Traverso]{ghallab-et-al-2004}
Malik Ghallab, Dana Nau, and Paolo Traverso.
\newblock \emph{Automated Planning: {Theory} and Practice}.
\newblock Morgan Kaufmann, 2004.

\bibitem[Guan et~al.(2023)Guan, Valmeekam, Sreedharan, and
  Kambhampati]{guan-et-al-neurips2023}
Lin Guan, Karthik Valmeekam, Sarath Sreedharan, and Subbarao Kambhampati.
\newblock Leveraging pre-trained large language models to construct and utilize
  world models for model-based task planning.
\newblock In \emph{Proc.\ NeurIPS 2023}, pp.\  79081--79094, 2023.

\bibitem[Hart et~al.(1968)Hart, Nilsson, and Raphael]{hart-et-al-ieeessc1968}
Peter~E. Hart, Nils~J. Nilsson, and Bertram Raphael.
\newblock A formal basis for the heuristic determination of minimum cost paths.
\newblock \emph{IEEE Transactions on Systems Science and Cybernetics},
  4\penalty0 (2):\penalty0 100--107, 1968.

\bibitem[Haslum et~al.(2007)Haslum, Botea, Helmert, Bonet, and
  Koenig]{haslum-et-al-aaai2007}
Patrik Haslum, Adi Botea, Malte Helmert, Blai Bonet, and Sven Koenig.
\newblock Domain-independent construction of pattern database heuristics for
  cost-optimal planning.
\newblock In \emph{Proc.\ {AAAI} 2007}, pp.\  1007--1012, 2007.

\bibitem[Haslum et~al.(2019)Haslum, Lipovetzky, Magazzeni, and
  Muise]{haslum-et-al-2019}
Patrik Haslum, Nir Lipovetzky, Daniele Magazzeni, and Christian Muise.
\newblock \emph{An Introduction to the Planning Domain Definition Language},
  volume~13 of \emph{Synthesis Lectures on Artificial Intelligence and Machine
  Learning}.
\newblock Morgan \& Claypool, 2019.

\bibitem[Helmert(2006)]{helmert-jair2006}
Malte Helmert.
\newblock The {Fast} {Downward} planning system.
\newblock \emph{Journal of Artificial Intelligence Research}, 26:\penalty0
  191--246, 2006.

\bibitem[Helmert(2009)]{helmert-aij2009}
Malte Helmert.
\newblock Concise finite-domain representations for {PDDL} planning tasks.
\newblock \emph{Artificial Intelligence}, 173:\penalty0 503--535, 2009.

\bibitem[Helmert \& Domshlak(2009)Helmert and
  Domshlak]{helmert-domshlak-icaps2009}
Malte Helmert and Carmel Domshlak.
\newblock Landmarks, critical paths and abstractions: {What's} the difference
  anyway?
\newblock In \emph{Proc.\ ICAPS 2009}, pp.\  162--169, 2009.

\bibitem[Helmert et~al.(2014)Helmert, Haslum, Hoffmann, and
  Nissim]{helmert-et-al-jacm2014}
Malte Helmert, Patrik Haslum, {J\"org} Hoffmann, and Raz Nissim.
\newblock Merge-and-shrink abstraction: A method for generating lower bounds in
  factored state spaces.
\newblock \emph{Journal of the ACM}, 61\penalty0 (3):\penalty0 16:1--63, 2014.

\bibitem[Hoffmann \& Nebel(2001)Hoffmann and Nebel]{hoffmann-nebel-jair2001}
J{\"o}rg Hoffmann and Bernhard Nebel.
\newblock The {FF} planning system: {Fast} plan generation through heuristic
  search.
\newblock \emph{Journal of Artificial Intelligence Research}, 14:\penalty0
  253--302, 2001.

\bibitem[Howey \& Long(2003)Howey and Long]{howey-long-icaps2003wscompetition}
Richard Howey and Derek Long.
\newblock {VAL's} progress: The automatic validation tool for {PDDL2.1} used in
  the {International} {Planning} {Competition}.
\newblock In \emph{ICAPS 2003 Workshop on the Competition}, 2003.

\bibitem[Hubert et~al.(2026)Hubert, Mehta, Sartran, Horv{\'a}th,
  {\v{Z}}u{\v{z}}i{\'c}, Wieser, Huang, Schrittwieser, Schroecker, Masoom,
  Bertolli, Zahavy, Mandhane, Yung, Beloshapka, Ibarz, Veeriah, Yu, Nash,
  Lezeau, Mercuri, S{\"o}nne, Mehta, Davies, Zheng, Pedregosa, Li, von Glehn,
  Rowland, Albanie, Velingker, Schmitt, Lockhart, Hughes, Michalewski,
  Sonnerat, Hassabis, Kohli, and Silver]{hubert-et-al-nature2026}
Thomas Hubert, Rishi Mehta, Laurent Sartran, Mikl{\'o}s~Z. Horv{\'a}th, Goran
  {\v{Z}}u{\v{z}}i{\'c}, Eric Wieser, Aja Huang, Julian Schrittwieser, Yannick
  Schroecker, Hussain Masoom, Ottavia Bertolli, Tom Zahavy, Amol Mandhane,
  Jessica Yung, Iuliya Beloshapka, Borja Ibarz, Vivek Veeriah, Lei Yu, Oliver
  Nash, Paul Lezeau, Salvatore Mercuri, Calle S{\"o}nne, Bhavik Mehta, Alex
  Davies, Daniel Zheng, Fabian Pedregosa, Yin Li, Ingrid von Glehn, Mark
  Rowland, Samuel Albanie, Ameya Velingker, Simon Schmitt, Edward Lockhart,
  Edward Hughes, Henryk Michalewski, Nicolas Sonnerat, Demis Hassabis, Pushmeet
  Kohli, and David Silver.
\newblock Olympiad-level formal mathematical reasoning with reinforcement
  learning.
\newblock \emph{Nature}, 651\penalty0 (8106):\penalty0 607--613, 2026.

\bibitem[Jiang et~al.(2023)Jiang, Welleck, Zhou, Li, Liu, Jamnik, Lacroix,
  Lample, and Wu]{jiang-et-al-iclr2023}
Albert~Q. Jiang, Sean Welleck, Jin~Peng Zhou, Wenda Li, Jiacheng Liu, Mateja
  Jamnik, Timoth{\'e}e Lacroix, Guillaume Lample, and Yuhuai Wu.
\newblock Draft, sketch, and prove: Guiding formal theorem provers with
  informal proofs.
\newblock In \emph{Proc.\ ICLR 2023}, 2023.

\bibitem[Junghanns \& Schaeffer(2001)Junghanns and
  Schaeffer]{junghanns-schaeffer-aij2001}
Andreas Junghanns and Jonathan Schaeffer.
\newblock Sokoban: {Enhancing} general single-agent search methods using domain
  knowledge.
\newblock \emph{Artificial Intelligence}, 129\penalty0 (1--2):\penalty0
  219--251, 2001.

\bibitem[Kambhampati et~al.(2024)Kambhampati, Valmeekam, Guan, Verma, Stechly,
  Bhambri, Saldyt, and Murthy]{kambhampati-et-al-icml2024}
Subbarao Kambhampati, Karthik Valmeekam, Lin Guan, Mudit Verma, Kaya Stechly,
  Siddhant Bhambri, Lucas~Paul Saldyt, and Anil~B. Murthy.
\newblock Position: {LLMs} can't plan, but can help planning in {LLM-Modulo}
  frameworks.
\newblock In \emph{Proc.\ ICML 2024}, pp.\  22895--22907, 2024.

\bibitem[Katz et~al.(2024)Katz, Kokel, Srinivas, and
  Sohrabi]{katz-et-al-neurips2024}
Michael Katz, Harsha Kokel, Kavitha Srinivas, and Shirin Sohrabi.
\newblock Thought of search: Planning with language models through the lens of
  efficiency.
\newblock In \emph{Proc.\ NeurIPS 2024}, pp.\  138491--138568, 2024.

\bibitem[Li et~al.(2026)Li, Yang, Song, Li, and Yang]{li-et-al-arxiv2026p3}
Zenan Li, Ziran Yang, Peiyang Song, Zhaoyu Li, and Kaiyu Yang.
\newblock {$P^3$}: Joint program-and-proof planning for verified code
  generation.
\newblock arXiv:2608.09277 [cs.AI], 2026.

\bibitem[Liu et~al.(2024)Liu, Tong, Yuan, Lin, Luo, Wang, Lu, and
  Zhang]{liu-et-al-icml2024}
Fei Liu, Xialiang Tong, Mingxuan Yuan, Xi~Lin, Fu~Luo, Zhenkun Wang, Zhichao
  Lu, and Qingfu Zhang.
\newblock Evolution of heuristics: Towards efficient automatic algorithm design
  using large language model.
\newblock In \emph{Proc.\ ICML 2024}, pp.\  32201--32223, 2024.

\bibitem[Marom \& Rosman(2020)Marom and Rosman]{marom-rosman-icaps2020}
Ofir Marom and Benjamin Rosman.
\newblock Utilising uncertainty for efficient learning of likely-admissible
  heuristics.
\newblock In \emph{Proc.\ ICAPS 2020}, pp.\  560--568, 2020.

\bibitem[McConnell et~al.(2011)McConnell, Mehlhorn, N{\"a}her, and
  Schweitzer]{mcconnell-et-al-csr2011}
Ross~M. McConnell, Kurt Mehlhorn, Stefan N{\"a}her, and Pascal Schweitzer.
\newblock Certifying algorithms.
\newblock \emph{Computer Science Review}, 5\penalty0 (2):\penalty0 119--161,
  2011.

\bibitem[Mugdan et~al.(2023)Mugdan, Christen, and
  Eriksson]{mugdan-et-al-icaps2023}
Esther Mugdan, Remo Christen, and Salom{\'e} Eriksson.
\newblock Optimality certificates for classical planning.
\newblock In \emph{Proc.\ ICAPS 2023}, pp.\  286--294, 2023.

\bibitem[Murray et~al.(2026)Murray, Dervovic, Pozanco, and
  Cashmore]{murray-et-al-icaps2026}
Andrew Murray, Danial Dervovic, Alberto Pozanco, and Michael Cashmore.
\newblock {GenePlan}: Evolving better generalized {PDDL} plans using large
  language models.
\newblock In \emph{Proc.\ ICAPS 2026}, pp.\  667--676, 2026.

\bibitem[Nipkow et~al.(2002)Nipkow, Wenzel, and Paulson]{nipkow-et-al-2002}
Tobias Nipkow, Markus Wenzel, and Lawrence~C. Paulson.
\newblock \emph{Isabelle/{HOL}: A Proof Assistant for Higher-Order Logic}.
\newblock Springer-Verlag, 2002.

\bibitem[Novikov et~al.(2025)Novikov, V{\~u}, Eisenberger, Dupont, Huang,
  Wagner, Shirobokov, Kozlovskii, Ruiz, Mehrabian, Kumar, See, Chaudhuri,
  Holland, Davies, Nowozin, Kohli, and Balog]{novikov-et-al-arxiv2025}
Alexander Novikov, Ng{\^a}n V{\~u}, Marvin Eisenberger, Emilien Dupont, Po-Sen
  Huang, Adam~Zsolt Wagner, Sergey Shirobokov, Borislav Kozlovskii, Francisco
  J.~R. Ruiz, Abbas Mehrabian, M.~Pawan Kumar, Abigail See, Swarat Chaudhuri,
  George Holland, Alex Davies, Sebastian Nowozin, Pushmeet Kohli, and Matej
  Balog.
\newblock {AlphaEvolve}: A coding agent for scientific and algorithmic
  discovery.
\newblock arXiv:2506.13131, 2025.

\bibitem[Pearl(1984)]{pearl-1984}
Judea Pearl.
\newblock \emph{Heuristics: {Intelligent} Search Strategies for Computer
  Problem Solving}.
\newblock Addison-Wesley, 1984.

\bibitem[Pereira et~al.(2026)Pereira, Corr{\^e}a, and
  Seipp]{pereira-et-al-neurips2026}
Andr{\'e}~Grahl Pereira, Augusto~B. Corr{\^e}a, and Jendrik Seipp.
\newblock Property-guided {LLM} program synthesis for planning.
\newblock In \emph{Proc.\ NeurIPS 2026}, 2026.

\bibitem[Phung et~al.(2026)Phung, Drexler, Lequen, and
  Seipp]{phung-et-al-icaps2026wsripl}
Windy Phung, Dominik Drexler, Arnaud Lequen, and Jendrik Seipp.
\newblock {LLM}-evolved pattern generators for optimal classical planning.
\newblock In \emph{ICAPS Workshop on Reliability In Planning and Learning
  (RIPL)}, 2026.

\bibitem[Ren et~al.(2025)Ren, Shao, Song, Xin, Wang, Zhao, Zhang, Fu, Zhu,
  Yang, Wu, Gou, Ma, Tang, Liu, Gao, Guo, and Ruan]{ren-et-al-arxiv2025}
Z.~Z. Ren, Zhihong Shao, Junxiao Song, Huajian Xin, Haocheng Wang, Wanjia Zhao,
  Liyue Zhang, Zhe Fu, Qihao Zhu, Dejian Yang, Z.~F. Wu, Zhibin Gou, Shirong
  Ma, Hongxuan Tang, Yuxuan Liu, Wenjun Gao, Daya Guo, and Chong Ruan.
\newblock {DeepSeek-Prover-V2}: Advancing formal mathematical reasoning via
  reinforcement learning for subgoal decomposition.
\newblock arXiv:2504.21801 [cs.CL], 2025.

\bibitem[Romera-Paredes et~al.(2024)Romera-Paredes, Barekatain, Novikov, Balog,
  Kumar, Dupont, Ruiz, Ellenberg, Wang, Fawzi, Kohli, and
  Fawzi]{romera-paredes-et-al-nature2024}
Bernardino Romera-Paredes, Mohammadamin Barekatain, Alexander Novikov, Matej
  Balog, M.~Pawan Kumar, Emilien Dupont, Francisco J.~R. Ruiz, Jordan~S.
  Ellenberg, Pengming Wang, Omar Fawzi, Pushmeet Kohli, and Alhussein Fawzi.
\newblock Mathematical discoveries from program search with large language
  models.
\newblock \emph{Nature}, 625:\penalty0 468--475, 2024.

\bibitem[Seipp(2024)]{seipp-ecai2024}
Jendrik Seipp.
\newblock Dissecting {Scorpion}: Ablation study of an optimal classical
  planner.
\newblock In \emph{Proc.\ ECAI 2024}, pp.\  39--42, 2024.

\bibitem[Seipp et~al.(2017)Seipp, Pommerening, Sievers, and
  Helmert]{seipp-et-al-zenodo2017}
Jendrik Seipp, Florian Pommerening, Silvan Sievers, and Malte Helmert.
\newblock {Downward} {Lab}.
\newblock \url{https://doi.org/10.5281/zenodo.790461}, 2017.

\bibitem[Seipp et~al.(2020)Seipp, Keller, and Helmert]{seipp-et-al-jair2020}
Jendrik Seipp, Thomas Keller, and Malte Helmert.
\newblock Saturated cost partitioning for optimal classical planning.
\newblock \emph{Journal of Artificial Intelligence Research}, 67:\penalty0
  129--167, 2020.

\bibitem[Shen et~al.(2020)Shen, Trevizan, and
  Thi{\'e}baux]{shen-et-al-icaps2020}
William Shen, Felipe Trevizan, and Sylvie Thi{\'e}baux.
\newblock Learning domain-independent planning heuristics with hypergraph
  networks.
\newblock In \emph{Proc.\ ICAPS 2020}, pp.\  574--584, 2020.

\bibitem[Silver et~al.(2024)Silver, Dan, Srinivas, Tenenbaum, {Pack Kaelbling},
  and Katz]{silver-et-al-aaai2024}
Tom Silver, Soham Dan, Kavitha Srinivas, Joshua~B. Tenenbaum, Leslie {Pack
  Kaelbling}, and Michael Katz.
\newblock Generalized planning in {PDDL} domains with pretrained large language
  models.
\newblock In \emph{Proc.\ {AAAI} 2024}, pp.\  20256--20264, 2024.

\bibitem[Solar-Lezama et~al.(2006)Solar-Lezama, Tancau, Bod{\'\i}k, Seshia, and
  Saraswat]{solar-lezama-et-al-asplos2006}
Armando Solar-Lezama, Liviu Tancau, Rastislav Bod{\'\i}k, Sanjit~A. Seshia, and
  Vijay~A. Saraswat.
\newblock Combinatorial sketching for finite programs.
\newblock In \emph{Proceedings of the 12th International Conference on
  Architectural Support for Programming Languages and Operating Systems (ASPLOS
  2006)}, pp.\  404--415. {ACM}, 2006.

\bibitem[Stechly et~al.(2024)Stechly, Valmeekam, and
  Kambhampati]{stechly-et-al-neurips2024}
Kaya Stechly, Karthik Valmeekam, and Subbarao Kambhampati.
\newblock Chain of thoughtlessness? {An} analysis of {CoT} in planning.
\newblock In \emph{Proc.\ NeurIPS 2024}, pp.\  29106--29141, 2024.

\bibitem[Stein et~al.(2026)Stein, Hodel, Fi{\v{s}}er, Hoffmann, Katz, and
  Koller]{stein-et-al-icaps2026}
Katharina Stein, Nils Hodel, Daniel Fi{\v{s}}er, J{\"o}rg Hoffmann, Michael
  Katz, and Alexander Koller.
\newblock Improved generalized planning with {LLMs} through strategy refinement
  and reflection.
\newblock In \emph{Proc.\ ICAPS 2026}, pp.\  677--686, 2026.

\bibitem[Taitler et~al.(2024)Taitler, Alford, Espasa, Behnke, Fi{\v{s}}er,
  Gimelfarb, Pommerening, Sanner, Scala, Schreiber, Segovia-Aguas, and
  Seipp]{taitler-et-al-aimag2024}
Ayal Taitler, Ron Alford, Joan Espasa, Gregor Behnke, Daniel Fi{\v{s}}er,
  Michael Gimelfarb, Florian Pommerening, Scott Sanner, Enrico Scala, Dominik
  Schreiber, Javier Segovia-Aguas, and Jendrik Seipp.
\newblock The 2023 {International Planning Competition}.
\newblock \emph{AI Magazine}, 45\penalty0 (2):\penalty0 280--296, 2024.
\newblock \doi{10.1002/aaai.12169}.

\bibitem[Tantakoun et~al.(2025)Tantakoun, Muise, and
  Zhu]{tantakoun-et-al-acl2025}
Marcus Tantakoun, Christian Muise, and Xiaodan Zhu.
\newblock {LLM}s as planning formalizers: A survey for leveraging large
  language models to construct automated planning models.
\newblock In \emph{Findings of the Association for Computational Linguistics:
  {ACL} 2025}, pp.\  25167--25188. Association for Computational Linguistics,
  2025.

\bibitem[Tuisov et~al.(2026)Tuisov, Vernik, and
  Shleyfman]{tuisov-et-al-icaps2026}
Alexander Tuisov, Yonatan Vernik, and Alexander Shleyfman.
\newblock Successor-generator planning with {LLM}-generated heuristics.
\newblock In \emph{Proc.\ ICAPS 2026}, pp.\  332--341, 2026.

\bibitem[Valmeekam et~al.(2023)Valmeekam, Marquez, Sreedharan, and
  Kambhampati]{valmeekam-et-al-neurips2023}
Karthik Valmeekam, Matthew Marquez, Sarath Sreedharan, and Subbarao
  Kambhampati.
\newblock On the planning abilities of large language models - {A} critical
  investigation.
\newblock In \emph{Proc.\ NeurIPS 2023}, pp.\  75993--76005, 2023.

\bibitem[Xin et~al.(2024)Xin, Guo, Shao, Ren, Zhu, Liu, Ruan, Li, and
  Liang]{xin-et-al-arxiv2024}
Huajian Xin, Daya Guo, Zhihong Shao, Zhizhou Ren, Qihao Zhu, Bo~Liu, Chong
  Ruan, Wenda Li, and Xiaodan Liang.
\newblock {DeepSeek-Prover}: Advancing theorem proving in {LLMs} through
  large-scale synthetic data.
\newblock arXiv:2405.14333 [cs.AI], 2024.

\bibitem[Yang et~al.(2023)Yang, Swope, Gu, Chalamala, Song, Yu, Godil, Prenger,
  and Anandkumar]{yang-et-al-neurips2023}
Kaiyu Yang, Aidan Swope, Alex Gu, Rahul Chalamala, Peiyang Song, Shixing Yu,
  Saad Godil, Ryan~J. Prenger, and Animashree Anandkumar.
\newblock {LeanDojo}: Theorem proving with retrieval-augmented language models.
\newblock In \emph{Proc.\ NeurIPS 2023}, pp.\  21573--21612, 2023.

\bibitem[Ye et~al.(2024)Ye, Wang, Cao, Berto, Hua, Kim, Park, and
  Song]{ye-et-al-neurips2024}
Haoran Ye, Jiarui Wang, Zhiguang Cao, Federico Berto, Chuanbo Hua, Haeyeon Kim,
  Jinkyoo Park, and Guojie Song.
\newblock {ReEvo}: Large language models as hyper-heuristics with reflective
  evolution.
\newblock In \emph{Proc.\ NeurIPS 2024}, pp.\  43571--43608, 2024.

\bibitem[Ye et~al.(2026)Ye, Yan, He, Kasriel, Yang, and
  Song]{ye-et-al-iclr2026}
Zhe Ye, Zhengxu Yan, Jingxuan He, Timothe Kasriel, Kaiyu Yang, and Dawn Song.
\newblock {VERINA}: Benchmarking verifiable code generation.
\newblock In \emph{Proc.\ ICLR 2026}, 2026.

\end{thebibliography}
